\documentclass{article}

\usepackage{microtype}
\usepackage{graphicx}
\usepackage{subcaption}
\usepackage{booktabs} % for professional tables

\usepackage{hyperref}

\usepackage[preprint]{icml2026}

\usepackage{amsmath}
\usepackage{amssymb}
\usepackage{mathtools}
\usepackage{amsthm}

\usepackage{csquotes}
\usepackage{caption}
\usepackage{wrapfig}
\usepackage{subcaption}
\usepackage{xcolor}
\usepackage{bbm}

\usepackage{enumitem}

\newcommand{\mD}{\mathbf{D}}

\newcommand{\mI}{\mathbf{I}}

\newcommand{\bh}{\mathbf{h}} 
\newcommand{\bx}{\mathbf{x}}
\newcommand{\by}{\mathbf{y}}
\newcommand{\bg}{\mathbf{g}}
\newcommand{\bJ}{\mathbf{J}}
\newcommand{\bD}{\mathbf{D}}
\newcommand{\bI}{\mathbf{I}}

\newcommand{\cL}{\mathcal{L}}
\newcommand{\cU}{\mathcal{U}}
\newcommand{\cS}{\mathcal{S}}

\usepackage[capitalize,noabbrev]{cleveref}

\theoremstyle{plain}

\theoremstyle{definition}

\theoremstyle{remark}

\usepackage[textsize=tiny]{todonotes}

\icmltitlerunning{Efficient Sparse Selective-Update RNNs for Long-Range Sequence Modeling}

\begin{document}

\twocolumn[
  \icmltitle{Efficient Sparse Selective-Update RNNs for Long-Range Sequence Modeling}

  \icmlsetsymbol{equal}{*}

  \begin{icmlauthorlist}
    \icmlauthor{Bojian Yin}{ia}
    \icmlauthor{Shurong Wang}{ia,zd}
    \icmlauthor{Haoyu Tan}{tue}
    \icmlauthor{Sander Bohte}{cwi}
    \icmlauthor{Federico Corradi}{tue}
    \icmlauthor{Guoqi Li}{ia}
  \end{icmlauthorlist}

  \icmlaffiliation{ia}{Institute of Automation, Chinese Academy of Sciences}
  \icmlaffiliation{tue}{Eindhoven University of Technology}
  \icmlaffiliation{zd}{Zhejiang University}
  \icmlaffiliation{cwi}{Machine Learning Group, Centrum Wiskunde \& Informatica (CWI)}

  \icmlcorrespondingauthor{Bojian Yin}{bojian.yin@ia.ac.cn}
  \icmlcorrespondingauthor{Guoqi Li}{guoqi.li@ia.ac.cn}

  \icmlkeywords{Machine Learning, ICML}

  \vskip 0.3in
]

\printAffiliationsAndNotice{}  % suppress ICML proceedings notice

\begin{abstract}
Real-world sequential signals, such as audio or video, contain critical information that is often embedded within long periods of silence or noise. 
While recurrent neural networks (RNNs) are designed to process such data efficiently, they often suffer from ``memory decay'' due to a rigid update schedule: they typically update their internal state at every time step, even when the input is static. 
This constant activity forces the model to overwrite its own memory and makes it hard for the learning signal to reach back to distant past events.
Here we show that we can overcome this limitation using Selective-Update RNNs (suRNNs), a non-linear architecture that learns to preserve its memory when the input is redundant. 
By using a neuron-level binary switch that only opens for informative events, suRNNs decouple the recurrent updates from the raw sequence length.
This mechanism allows the model to maintain an exact, unchanged memory of the past during low-information intervals, creating a direct path for gradients to flow across time. 
Our experiments on the Long Range Arena, WikiText, and other synthetic benchmarks show that suRNNs match or exceed the accuracy of much more complex models such as Transformers, while remaining significantly more efficient for long-term storage. 
By allowing each neuron to learn its own update timescale, our approach resolves the mismatch between how long a sequence is and how much information it actually contains. 
By providing a principled approach to managing temporal information density, this work establishes a new direction for achieving Transformer-level performance within the highly efficient framework of recurrent modeling.
\end{abstract}

\begin{figure*}[!ht]
    \centering
    \includegraphics[scale = 1.0]{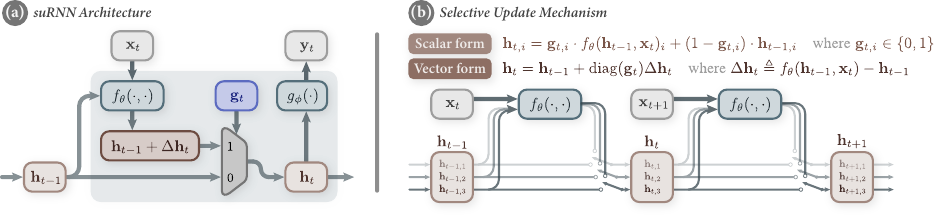}
    
    \caption{\textbf{suRNN architecture and the selective update mechanism}: 
    (a) Transition of suRNN from time step $t-1$ to $t$: unlike conventional RNNs that apply a uniform, time-agnostic transition, suRNN adopts a per-neuron, time-dependent binary gate $\bg_t \in \{0, 1\}^H$.
    This allows each neuron to dynamically choose whether to preserve or update its state at each step.
    (b) Selective-update mechanism for $H = 3$ neurons: the per-neuron gating logic functions as an RC circuit with a switch.
    When the switch is off ($\bg_{t, i} = 0$, bottom path), neuron $\bh_{t, i}$ bypasses the update and remains unchanged;
    when the switch is on ($\bg_{t, i} = 1$, top path), neuron $\bh_{t, i}$ undergoes a standard non-linear update.
    }

    \label{fig:1_summary}
\end{figure*}

\section{Introduction}

Recurrent Neural Networks (RNNs) are a classical and foundational framework for sequential modeling \cite{elman1990rnn1, hopfield1982rnn2, mcculloch1943rnn3}.
Theoretically, RNNs are capable of representing arbitrary state-dependent computations through Turing-complete recurrence \cite{kilian1996rnn5, keiblinger2025recurrence}.
Practically, RNNs are attractive for $\mathcal O(1)$ memory complexity during autoregressive inference and low-latency decoding, making them natural candidates for streaming, on-device, and ultra-long sequence deployment \cite{graves2012rnn6,liu2019rnn4}.
Despite this elegant formulation and favorable inference efficiency, RNNs have been largely surpassed by Transformers \cite{vaswani2017attention} and modern State Space Models (SSMs) \cite{Gu2021S4, Tay2022EfficientSurvey}, particularly in tasks where capturing long-range dependencies is essential.
Real-world sequential data (e.g., text, audio, or video) exhibit highly non-uniform information density: significant events are typically sparse and separated by long intervals of redundancy or noise.
Ideally, a sequence model should allocate computation effort in proportion to information density rather than raw sequence length.
However, many widely used models remain \textit{information-agnostic} in this regard.
For instance, Transformers compute pairwise interactions across time steps via $\mathcal O(L^2)$ attention map, while SSMs integrate history through long-range convolutions.
Yet, these models have access to distant context but they process each time step uniformly, expending equal effort on informative and redundant content.
RNNs face an even worse situation where they apply a dense, uniform, and \textit{time-agnostic} transformation at every time step, forcing the valuable historical information to be constantly overwritten and re-encoded.
This synchronization of computation with sequence length prevents the model from distinguishing meaningful transitions from non-informative ones, resulting in unnecessary state updates that are wasteful and can actively degrade long-term retention.

To bridge this mismatch, we propose \textbf{Selective-Update RNNs (suRNNs)} (Fig.~\ref{fig:1_summary}a), an architecture designed to alleviate the long-term dependency problem through temporal sparsity at the neuron level.
Our core insight is that stable long-sequence retention requires two properties:
(i) an identity map to preserve states during non-informative intervals, and (ii) a mechanism to perform informative updates selectively.
suRNNs instantiate these by replacing conventional continuous gating with a time-dependent, neuron-wise binary gate $\bg_{t,i} \in \{0, 1\}$, thereby decoupling effective recurrent updates from raw sequence.
This modification directly overcomes the time-agnostic update schedule of conventional RNNs, and injects the temporal awareness to the model through diverse gating schedules. 
To handle the non-differentiability of the discrete gates, we adopt the Straight-Through Estimator (STE) \cite{bengio2013ste}, drawing a direct parallel to surrogate gradient methods in Spiking Neural Networks \cite{neftci2019surrogate}.

As depicted in Fig.~\ref{fig:1_summary}b, selective update induces a dual-mode dynamics.
When the gate is switched off ($\bg_{t,i} = 0$), the $i$-th neuron acts as an ideal memory cell, preserving the exact same state from the previous time step;
when the gate is switched on ($\bg_{t,i} = 1$) for the $i$-th neuron, a standard nonlinear update is applied. 
Consequently, this mechanism induces a \textit{sparse credit-assignment} effect, in which the number of non-identity transitions along a gradient path scales with the number of informative events (gate activations) rather than the raw sequence length.
% By mathematically masking parameter updates during redundant intervals, suRNNs focus the learning signal on salient transitions.
% Selective updates induce a sparse credit assignment mechanism where the effective gradient path for each neuron scales with the number of informative events rather than raw sequence length, mathematically masking parameter updates during redundant timesteps to focus the learning signal on salient transitions.
% As the consequence, the backward gradients propagate by the number of informative updates rather than raw sequence length. 
% Although BPTT still unrolls for T steps, identity carries make the non-identity factors in the gradient product scale with the number of informative updates rather than  raw sequence length.

% Unlike prior adaptive computation methods such as Adaptive Computation Time (ACT) \cite{graves2016adaptive}, Skip RNNs \cite{campos2018skip} or by imposing structured or stochastic schedules \cite{Neil2016PhasedLSTM}, which often require complex ponder losses or global halting logic, suRNNs implement neuron-level sparsity within the standard recurrence. This allows for heterogeneous effective timescales across the hidden state, concentrating computation only on salient moments. Our main contributions are as follows:

Unlike prior adaptive computation approaches --- such as the global halting logic of Adaptive Computation Time \cite{graves2016adaptive}, the discrete step-skipping of Skip RNNs \cite{campos2018skip}, or the rigid schedules of Clockwork RNNs \cite{Koutnik2014Clockwork} --- suRNNs implement sparsity at the finest possible resolution: the individual neuron.
These earlier methods often struggle to scale to larger models and longer sequences, as they are hindered by the optimization instability of auxiliary ponder losses or the sequential bottlenecks of global control flow \cite{campos2018skip}.
By contrast, suRNNs integrate sparsity directly into the recurrent updates, enabling heterogeneous effective timescales to emerge naturally across the hidden state.
This per-neuron selectivity avoids global contextual synchronization on all neurons, ensures stable training on large-scale datasets, and concentrates representational capacity exclusively on salient transitions.
Our main contributions are as follows:
\begin{itemize}%[noitemsep, topsep=0pt]
    \item \textbf{Selective Update Mechanism}: We introduce suRNN, an architecture that replaces continuous gating with  binary selection to enable exact state preservation during intervals of informational stasis.
    \item \textbf{Sparse Credit Assignment}: We apply the STE to recurrent gating, inducing a functional gradient path that scales with the number of salient events rather than raw sequence length, effectively bypassing the vanishing/explosion gradient problem.
    \item \textbf{Strong Empirical Performance}: We show that suRNN outperforms existing RNNs in streaming tasks such as psMNIST, sCIFAR, and selective copy. Furthermore, suRNN matches or exceeds the accuracy of modern Transformers on the Long Range Arena (LRA) and WikiText benchmarks while maintaining $\mathcal O(1)$ inference efficiency.
\end{itemize}

\section{Background}

% \subsection{Standard RNNs and BPTT}

\subsection{Recurrent Neural Network}
\label{sec:method_rnn}

Conceptually, recurrent neural networks (RNNs) map an input sequence $\{ \mathbf{x}_t \}_{t=1}^T$ to hidden states $\{ \mathbf{h}_t \}_{t=1}^T$, where $\mathbf{h}_t \in \mathbb{R}^H$ summarizes the history up to time $t$.
A standard recurrent layer evolves according to:
\begin{equation}
\label{eq:std_rnn}
\bh_t = f_{\theta}(\bh_{t-1}, \bx_t), \quad\text{for}\; 1 \le t \le T,
\end{equation}
where $f_\theta(\cdot, \cdot)$ is parameterized by $\theta$ and generates the next state $\bh_t$ from the previous state $\bh_{t-1}$ and input $\bx_t$.
At each time step $t$, a readout $\by_t = g_\phi(\bh_t)$ is computed from the current hidden state $\bh_t$, and a step-wise loss $\ell_t = \ell(\by_t,\by_t^\ast)$ measures the disagreement between $\by_t$ and the ground truth $\by_t^\ast$.
An overall loss $\cL(\theta, \phi)$ can then be defined to measure the discrepancy between the model outputs $\{ \by_t \}_{t=1}^T$ and the targets $\{ \by^\ast_t \}_{t=1}^T$:
\begin{equation}
\cL(\theta, \phi) = \sum_{t=1}^T \ell_t = \sum_{t=1}^T \ell\big( g_\phi(f_\theta(\bh_{t-1}, \bx_t)), \by^\ast_t \big)
\end{equation}

Standard RNNs are time-agnostic in the sense that the same transition function $f_\theta$ is applied at every time step.
Despite being parameter-efficient, this design can be restrictive under non-stationary or time-dependent dynamics, where the appropriate update rule should vary over time.

\subsection{Backpropagation Through Time}

Training RNNs requires minimizing the overall loss $\cL(\theta,\phi)$.
Since the loss $\cL$ involves the iterative states $\{ \bh_t \}_{t=1}^{T}$, it is typically optimized via BPTT, which unrolls hidden states backward in time:
\begin{equation}
\label{eq:J_rnn}
\frac{\partial \cL}{\partial \bh_{t-1}} = \frac{\partial \cL}{\partial \bh_t} \frac{\partial \bh_t}{\partial \bh_{t-1}} + \frac{\partial \ell_t}{\partial \bh_{t-1}}
\end{equation}
% Let $\mathbf{d h}_t := \partial \mathcal{L}/\partial \mathbf{h}_t$ denote the adjoint at time $t$. The backward recursion is
% \begin{equation}
% \label{eq:J_rnn}
% \mathbf{d h}_{t-1} = \frac{\partial \ell_{t-1}}{\partial \mathbf{h}_{t-1}}+\mathbf{d h}_t\, J_f^{t},
% \end{equation}
% where $J_f^{t} := \frac{\partial f(\mathbf{x}_t,\mathbf{h}_{t-1};\theta)}{\partial \mathbf{h}_{t-1}} \in \mathbb{R}^{H\times H}$.
Let $\bJ_t \triangleq \partial \bh_t / \partial \bh_{t-1} \in \mathbb R^{H \times H}$. Unrolling Eq.~\ref{eq:J_rnn} shows that credit assignment across $t{-}s$ steps involves the product $\prod_{\tau=s+1}^{t} \bJ_\tau$.
If $\|\bJ_\tau\| \le \rho$ holds uniformly, then $\big\| \prod_{\tau=s+1}^{t} \bJ_\tau \big\| \le \rho^{t-s}$, which implies that gradient vanishes when $\rho < 1$ and explodes when $\rho > 1$.

Therefore, the effective credit-assignment depth scales with sequence length, motivating architectures that reduce the multiplicative depth of backpropagation without changing the forward recurrence.

\section{Methods}

To address the above optimization issues, we propose Selective-Update RNNs (suRNNs), which replace continuous gating with a neuron-level binary mechanism. This shift enables exact identity mappings so that the hidden state can skip redundant updates and preserve information across long time spans.

\subsection{Selective Update}
\label{sec:method_su}

Building on the standard recurrent update in Section~\ref{sec:method_rnn}, we introduce a per-neuron binary gate $\bg_t \in \{0,1\}^H$ with diagonal mask $\bD_t = \operatorname{diag}(\bg_t)$ and reparameterize the state evolution of the recurrent network as:
\begin{equation}
\label{eq:su_update}
\begin{aligned}
\bh_t &= \bh_{t-1} + \bD_t\ \Delta \bh_t \\
&= \bh_{t-1} + \bD_t \big( f_\theta(\bh_{t-1}, \bx_t) - \bh_{t-1} \big) \\
&= (\mI - \bD_t) \, \bh_{t-1} + \bD_t \, f_\theta(\bh_{t-1}, \bx_t).
\end{aligned}
\end{equation}
Here, $\Delta \bh_t \in \mathbb{R}^H$ is the residual proposal, the deviation from identity produced by the original recurrent transformation, $\Delta \bh_t \triangleq f_\theta(\bh_{t-1}, \bx_t) - \bh_{t-1}$.
When $\bg_{t, i} = 0$, the $i$-th neuron (i.e., $\bh_{t, i}$) is carried exactly, while the usual recurrent step is applied when $\bg_{t,i} = 1$.
This mechanism preserves the update rule $f_\theta$ and its parameters and controls only \emph{when} $f_\theta$ is applied to each neuron.
In a mask-aware implementation, neurons with $\bg_{t,i} = 0$ need not contribute to $\Delta \bh_t$, so the cost of the forward and backward passes can scale with $\sum_{i=1}^H \bg_{t,i} \le H$ rather than $H$.
In practice, selective update can accelerate inference when the learned gates are sufficiently sparse and the implementation exploits that sparsity.

\begin{table*}[t]
\centering
\small
\caption{\textbf{Most informative structural trade-offs of sequence layers.}
In this table, $T$ is sequence length, $d$ is hidden size, $k$ is convolution kernel size, and $p\in[0,1]$ is the update rate (Eq.~\ref{eqn:upd_rate}). ``Compute / layer'' is worst-case training cost; ``Sequential ops'' counts inherently serial steps across time. ``Effective non-identity depth'' measures the typical number of non-identity transitions on a long-range route. ``Streaming state'' is memory required for strict online decoding, and ``AR inference / token'' is the per-token cost at context length $n$. The $\mathcal{O}(pn)$ and $\mathcal{O}(pd^{2})$ entries assume a mask-aware implementation that skips inactive neurons; otherwise selective-update recurrence matches dense RNN worst-case cost.}
\label{tab:layer_tradeoffs_compact}
\begin{tabular}{lccccc}
\toprule
\textbf{Layer type} &
\textbf{Compute / layer} &
\textbf{Sequential ops} &
\textbf{Effective depth} &
\textbf{Streaming state} &
\textbf{AR inference / token} \\
\midrule
Self Attention &
$\mathcal{O}(T^{2}d)$ &
$\mathcal{O}(1)$ &
$\mathcal{O}(1)$ &
$\mathcal{O}(Td)$ &
$\mathcal{O}(Td)$ \\
Conv$_k$ &
$\mathcal{O}(Tkd^{2})$ &
$\mathcal{O}(1)$ &
$\mathcal{O}(\log_{k} T)$ &
$\mathcal{O}(kd)$ &
$\mathcal{O}(kd^{2})$ \\
\midrule
RNN (dense) &
$\mathcal{O}(Td^{2})$ &
$\mathcal{O}(T)$ &
$\mathcal{O}(T)$ &
$\mathcal{O}(d)$ &
$\mathcal{O}(d^{2})$ \\
suRNN (ours) &
$\mathcal{O}(Td^{2})$ &
$\mathcal{O}(T)$ &
$\mathcal{O}(\overline p T)$ &
$\mathcal{O}(d)$ &
$\mathcal{O}(\overline p d^{2})$ \\
\bottomrule
\end{tabular}
% \vspace{2mm}

\end{table*}

Mathematically, linearizing the discrete map \eqref{eq:su_update} around $(\bx_t, \bh_{t-1})$ gives the one-step Jacobian:
\begin{equation}
\label{eq:su_jacobian}
\bJ_t \triangleq \frac{\partial \bh_t}{\partial \bh_{t-1}} = \bI + \mD_t \big( \bJ^{(f)}_t - \bI \big), 
\end{equation}
where $\bJ^{(f)}_t = \partial f_\theta / \partial \bh_{t-1} \, |_{(\bh_{t-1}, \bx_t)}$ denotes the Jacobian of $f_\theta$ (Eq.~\ref{eq:std_rnn}).
Eq.~\ref{eq:su_jacobian} mirrors the characteristic residual form $\bI + \partial f/\partial \bx$ as in ResNet \cite{He2016ResNet}, but now applied along the temporal axis.
The identity term provides an exact carry route for gradients, and the diagonal mask $\bD_t$ projects the residual correction onto the active neurons only.
In particular, if $\bg_{t,i} = 0$, the $i$-th row of $\bJ_t$ equals the corresponding row of $\bI$, so no change or mixing occurs for the $i$-th neuron at time $t$.
This row-wise identity property underpins the shorter gradient paths established in our analysis. 

% Conceptually,selective update makes the one-step Jacobian “identity plus masked residual,” which is the defining Jacobian form of residual connections, but applied along time rather than depth.

\subsection{Gate Scheduling}
\label{sec:method_gates}

We propose generating the gates $\bg$ with a \textit{rhythmic} module:
\begin{equation}
\label{eq:gate_fourier}
a_{t,i} = b_i + \sum_{k=1}^{K} \alpha_{ik} \sin(\omega_{k} \cdot t + \phi_{i,k}),  \;\;
\bg_{t,i} = \mathsf{H}(a_{t,i}).
\end{equation}
where $\mathsf{H}(\cdot)$ denotes the Heaviside step function. In this formulation, the $K$ frequencies $\{\omega_k\}$ are shared across the layer, while the amplitudes $\alpha_{ik}$, phases $\phi_{ik}$, and biases $b_i$ are learned per unit. By default, we set $K$ equal to the hidden dimension $H$.
% to provide sufficient representational capacity.

To facilitate end-to-end training through the non-differentiable Heaviside step function, we use a straight-through estimator \cite{bengio2013ste} or surrogate gradients \cite{neftci2019surrogate}. During the forward pass, the model uses the discrete gate $\bg_{t,i} = \mathsf{H}(a_{t,i}) = \mathbbm{1}\{a_{t,i} > 0\}$, while the backward pass uses a surrogate gradient derived from a sigmoid function $\sigma(x)$. 
% This allows the  parameters to receive a meaningful learning signal even though the actual update mechanism is discrete.
% Frequencies $\omega_{k}$ can be initialized on a linear or logarithmic grid. The sine table $\{\sin(\omega_k t)\}_{k=1}^K$ is precomputed per step, resulting in an $\mathcal O(KH)$ overhead.
% Purely data-driven gates (e.g., MLPs, SSMs) are drop-in alternatives; the theoretical analysis below does not depend on the sinusoidal choice.

For implementation efficiency, the frequencies $\omega_k$ are initialized on a logarithmic grid to span multiple orders of magnitude, and $\{\sin(\omega_k t)\}_{k=1}^K$ is precomputed, resulting in a marginal $\mathcal{O}(KH)$ computational overhead. Note that while this sinusoidal parameterization is effective for modeling structured intervals, the suRNN framework is agnostic to the gate generator; data-driven modules such as MLPs or SSMs can serve as drop-in alternatives without altering the underlying theoretical properties of the selective update.

\subsection{Selective update shortens effective gradient paths}
\label{sec:theory_su}

Let $1 \le s < t \le T$.
Following Eq.~\ref{eq:su_jacobian}, the Jacobian of the unrolled computation map from $\bh_s$ to $\bh_t$ is
\begin{equation}
\label{eq:su_sensitivity}
\frac{\partial \bh_t}{\partial \bh_s} = \prod_{\tau=s+1}^{t} \bJ_\tau
= \prod_{\tau=s+1}^{t} \big(\bI + \bD_\tau (\bJ^{(f)}_\tau- \bI) \big),
\end{equation}
which is exact for the forward trajectory $\{ (\bh_{\tau-1}, \bx_\tau)\}_{\tau=s}^{t}$.
For the $i$-th neuron, define the update set and count
\begin{align}
\label{eq:update_set}
\mathcal{U}(s, t)
& \triangleq \{ \tau \in \mathbb Z \mid s < \tau \le t \} \\
\mathcal{U}^{\text{on}}_i(s, t)
& \triangleq \{ \tau \in \mathcal U(s, t) \mid \bg_{\tau, i} = 1 \}
\end{align}

When $\bg_{\tau,i}=0$, the $i$-th row of $\bJ_\tau$ equals the $i$-th row of $\bI$ and would not expand the multiplicative chain for the $i$-th neuron.
Although BPTT still unrolls $T$ times, identity carries make the non-identity factors in the gradient product scale with the number of informative updates $|\mathcal U^{\text{on}}_i(s, t)|$ rather than sequence length $|\mathcal U(s, t)|$.

\paragraph{Assumptions and norms}
Let $\|\cdot\|$ denote any submultiplicative matrix norm (e.g., the operator norm induced by $\ell_2$ norm). Bounds below are stated for the \emph{row-wise} sensitivity $\big\| \partial \bh_{t,i} / \partial \bh_s \big\|$, interpreted via the induced norm on the $i$-th row of the Jacobian.

\paragraph{Proposition 1 (effective path length)}
\label{prop:effective_depth}
% Assume $\|\bJ_f^\tau\|\le \rho$ for all $\tau \in \mathcal U(s, t)$. For each neuron $i$ there exists a constant $C_i$, independent of $t{-}s$, such that
% \begin{equation}
% \label{eq:bound_coord}
% \Big\|\frac{\partial \bh_{t,i}}{\partial \bh_s}\Big\| \;\le\; C_i\, \rho^{|\mathcal U^{\text{on}}_i(s, t)|} < C_i \rho^{t - s}.
% \end{equation}
\label{prop:effective_path_length_corrected}
Assume that $\|J_\tau^{(f)}\|\le \rho$ for all $\tau\in U(s,t)$. Then, for each neuron $i$,
\begin{equation}
\left\|
e_i^\top \frac{\partial h_t}{\partial h_s}
\right\|
\le
C_i(s,t)\,\rho^{|U_i^{\mathrm{on}}(s,t)|},
\end{equation}
where $C_i(s,t)$ captures cross-coordinate coupling accumulated after update times. If these carry-only propagations are uniformly bounded, then $C_i(s,t)\le C_i$ for some constant $C_i$ independent of $t-s$.

In a conventional RNN that every neuron updates every step, the analogous bound scales as $\rho^{\,t-s}$. Thus the exponent that governs vanishing or exploding depends on the \emph{number of updates actually taken} by neuron $i$, not on elapsed steps. The constant $C_i$ absorbs cross-neuron coupling at update times (off-diagonal contributions in the $i$-th row when $g_{\tau,i}=1$) and does not grow with the sequence length.
\paragraph{Effective depth scales with update rate}
% If the average update rate on $[s,t]$ is $p_i:=\tfrac{1}{t-s}\sum_{\tau=s+1}^{t}\mathbbm{1}\{g_{\tau,i}=1\}$, then a typical effective depth scales like $p_i(t-s)$, and the bound becomes $C_i\,\rho^{\,p_i(t-s)}$.
Let $\overline p$ be the \textit{update rate}, which is the average number of times that the gate of each neuron is switched on for all time steps:
\begin{equation} \label{eqn:upd_rate}
    \overline p \triangleq \frac{1}{HT} \sum_{i=1}^H |\mathcal U^{\text{on}}_i(0, T)| \,<\, 1
\end{equation}
Then, the expected number of informative updates performed on each neuron is $\overline p \, T < T$, so the effective depth scales with update rate $\overline p$ rather than sequence length.
Plugging this into Proposition~1 gives $\|\partial \bh_T/\partial \bh_0\| \lesssim C\,\rho^{\,\overline p T}$, 
which makes explicit that reducing the update rate slows both gradient vanishing and exploding. 

Unlike the continuous gates in LSTMs \cite{Hochreiter1997LSTM} and GRUs \cite{Cho2014GRU} that (1) subject the hidden state to persistent fractional updates and force the temporal correlations to be asymptotically overwritten by transitions and (2) merely scales the spectral radius $\rho$, the binary gates in suRNN fundamentally reduce the effective gradient depth to $1 / \overline p$ times shorter than the raw sequence length.
This discretization transforms the optimization landscape from one of persistent decay to one of selective preservation.

%============================================================================
\paragraph{Backpropagation}
Combining Eq.~\ref{eq:J_rnn} with the selective-update Jacobian in Eq.~\ref{eq:su_jacobian}, the BPTT recursion becomes
\begin{equation}
\frac{\partial \cL}{\partial \bh_{t-1}} =
\frac{\partial \cL}{\partial \bh_t}\Big(
    \bI + \bD_t(\bJ^{(f)}_t - \bI)
\Big) +
\frac{\partial \ell_{t-1}}{\partial \bh_{t-1}} .
\end{equation}
non-update neurons propagate through an exact identity row: if $\bg_{t,i}=0$, the $i$-th row of $\bI + \bD_t(\bJ^{(f)}_t-\bI)$ equals the corr-esponding row of $\bI$, so no change and no mixing occur for neuron $i$ at time $t$. Consequently, parameter updates are sparse in time and in units:
\begin{equation}
\label{eq:param_grad_su}
\frac{\partial \cL}{\partial \theta}
=
\sum_{t=1}^{T}
\frac{\partial \cL}{\partial \bh_t}\,
\bD_t\,
\frac{\partial f_\theta(\bx_t,\bh_{t-1})}{\partial \theta},
\end{equation}
so, only active units at update steps contribute to $\partial \cL/\partial \theta$. Finally, when $\bg_{t,i}=\mathsf{H}(a_{t,i})$, the gradients reach the gate generator via a straight-through surrogate for $\mathsf{H}$, with local signal proportional to $\big(\frac{\partial \cL}{\partial \bh_{t,i}}\big)\Delta \bh_{t,i}$. 

\begin{figure}[htbp]
  \centering
  \begin{minipage}{0.5\textwidth}
    \includegraphics[width=\textwidth]{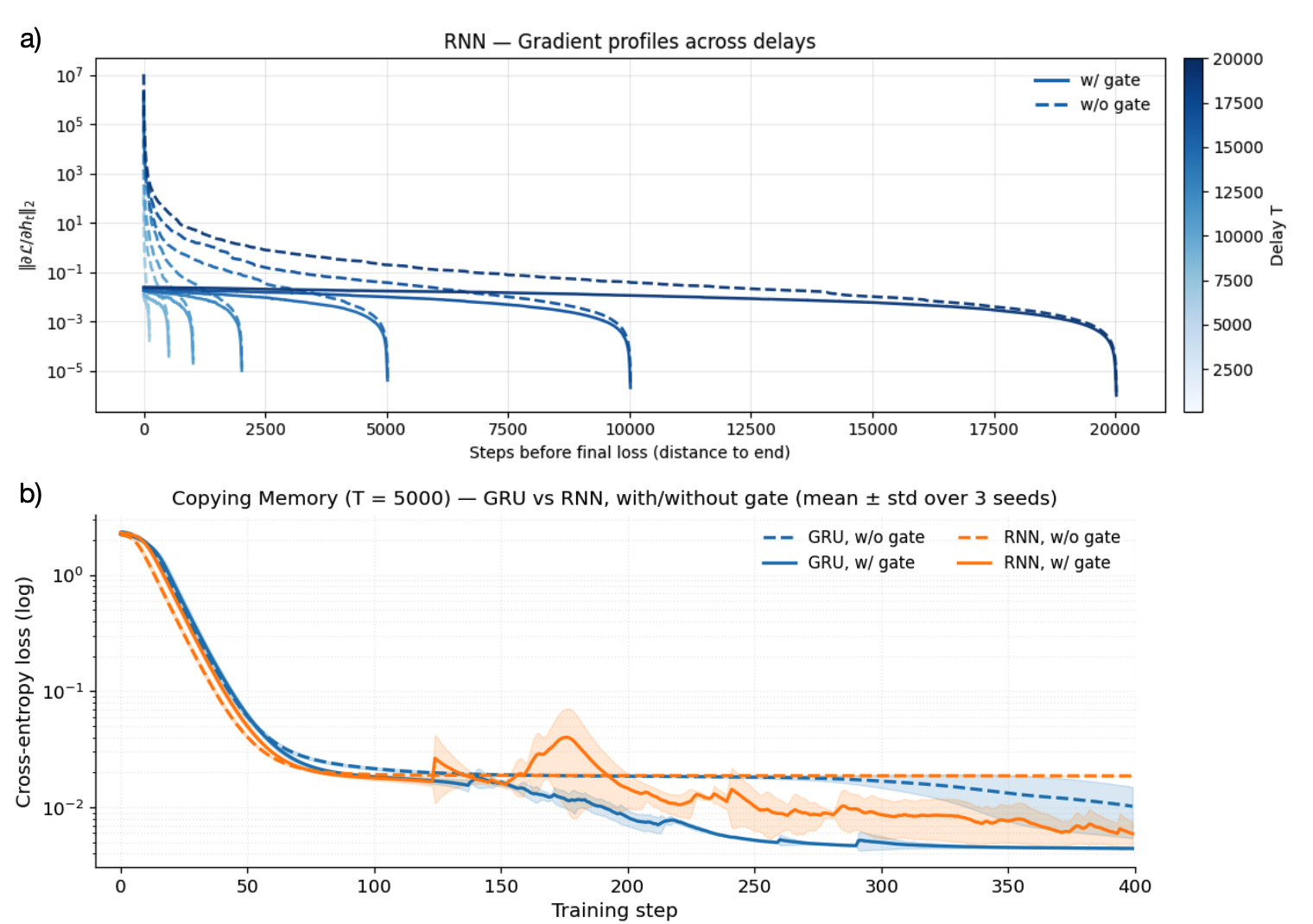}
  \end{minipage}%

  % \begin{minipage}{0.5\textwidth}
    \caption{\textbf{Selective update improves long-range credit assignment.}
    \textbf{(a)} RNN gradient profiles across delays $T$ (color bar; solid $=$ with selective gate, dashed $=$ without). Gating keeps gradients bounded (about $10^{-3}\!-\!10^{-2}$, log scale) with near-parallel decay; increasing $T$ mainly stretches the horizon, while ungated runs collapse to a tiny floor after early spikes. \textbf{(b)} Copying Memory at $T{=}5000$ (mean $\pm$ std over 3 seeds): both GRU and RNN with the gate converge faster and to lower loss, whereas ungated baselines plateau. These trends indicate the gate acts as an identity/skip path with per-step gain near $1$, preserving long-range gradients.}
        \label{fig:copy_task}
  % \end{minipage}
\end{figure}

For illustration, we revisit the classic Copying-Memory task \cite{le2015simple}, where the model observes a short prefix, then $T$ blanks and a delimiter, and must reproduce the prefix after the delimiter, stressing long-range dependency. For sequences with delay $T$, we track $\|\partial \ell/\partial \bh_\tau\|$ as a function of temporal distance for both RNN and GRU,  and their selective-update variants. With selective updates, in Fig. \ref{fig:copy_task} a), gradient traces remain bounded and nearly parallel as $T$ increases, consistent with the Jacobian structure $\bJ_t = \bI + \bD_t(\bJ^{(f)}_t - \bI)$ and Proposition~1, which predicts effective multiplicative depth governed by updates taken rather than input length.  On task with $T=5000$, the selective update gated models converge faster and reach a lower loss,
% indicating time-sparse credit concentrated on update steps and empirically validating the theoretical mechanism. 
and PCA trajectories (in Fig \ref{fig:copy_task_traj} in Appendix) show more structured state dynamics with clearer separation between delay and recall than a standard GRU.

\paragraph{Selective update creates ensembles of sub-RNNs}
\label{sec:expressivity}

% Let $\bU_t \triangleq \bD_t(\bJ^{(f)}_t - \bI)$.
% The product $\prod_{\tau=s+1}^{t} \bJ_\tau$ (Eq.~\ref{eq:su_sensitivity}) can be expanded to:
By expanding the product $\prod_{\tau=s+1}^{t} \bJ_\tau$ in Eq.~\ref{eq:su_sensitivity}:
\begin{equation}
\label{eq:ensemble}
% \mathbf I + \sum_{k=1}^{t-s}\ \sum_{s<\tau_1<\cdots<\tau_k\le t} \prod_{i=1}^k \mathbf U_{\tau_i},
\prod_{\tau=s+1}^t \big(\bI + \bD_\tau (\bJ^{(f)}_\tau - \bI)) =
\!\! \sum_{\cS \subseteq \cU(s, t)} \prod_{\tau_i \in \cS} \bD_{\tau_i} \big(\bJ^{(f)}_{\tau_i} - \bI \big),
\end{equation}

corresponding to distinct computation paths that apply the non-identity transform only on the selected timesteps and otherwise follow identity carries. On the computational graph, each particular gate realization therefore selects a sparse subgraph of the full unrolled RNN, yielding a sub-RNN that operates on a subsequence of updates; across inputs and layers, the model effectively integrates many such subgraphs \cite{graves2016adaptive}. This many-path view clarifies why selective updates can preserve functional depth while providing abundant short, identity-rich routes for gradient flow, paralleling the ensemble interpretation of residual networks~\cite{Veit2016EnsembleResNet}.

% move it to somewhere
In essence, selective updates increase usable memory , not memory capacity, by changing how information is preserved and how credit is assigned over time. 
It is enlarging the state by decoupling retention and credit assignment from un-selected timesteps. 
Non-updating neurons are carried exactly, avoiding repeated mixing and drift, and gradients traverse identity routes during these intervals. Consequently, effective temporal depth is governed by the number of selected subsequence rather than the whole sequence, yielding longer-lived representations and better-conditioned BPTT.

\subsection{Implementation}
Selective updates are a lightweight reparameterization and can be applied to essentially any recurrent backbone by inserting an identity-through-time carry with a learned temporal mask, without changing the underlying recurrent transform ( Fig~\ref{fig:rnn_summary} in Appendix). However, a direct stepwise realization still requires time consuming BPTT, which becomes a practical bottleneck as model size and context length grow. To enable scalable experiments, we introduce suGRU, a selective-update variant of the \textit{CUDA-fused} GRU that preserves execution speed, without any custom kernels or single-step control flow with minimal overhead. Implementation details are provided in Appendix~\ref{Apx:effsuGRU}.

% \vspace{-3pt}
% Beyond scalability, selective updates also offer potential inference acceleration and reduced memory traffic, since inactive neurons could skips forward recurrent computations and associated reads and writes can be also skipped under mask-aware execution.
\textbf{Efficiency} Beyond scalability, selective update improves both latency and memory traffic by skipping their recurrent computations and associated reads/writes under mask-aware execution.
While this conditional execution is not supported by default in PyTorch kernels, we use a mask-aware C implementation of stepwise suGRU with block gating on sequential MNIST, we measure a $5.3\times$ latency reduction at 83\% sparsity ($466$ ms $\rightarrow$ $88$ ms per step; details in the Appendix \ref{sec:time}). This event-triggered computation is naturally compatible with sparse-kernel, neuromorphic, and other event-driven hardware that supports conditional state updates.
\hfill\hfill\break

\begin{table*}[!t]
\centering
\small
\setlength{\tabcolsep}{5pt} % Reduced slightly to fit extra column
\caption{\textbf{Long Range Arena (LRA) Results.} We compare suGRU against Transformer variants and State Space Models (S4). The \textbf{Streaming?} column indicates strictly uni-directional, streaming processing. Reported results are taken from \cite{gu2022s4}}
\label{tab:LRA}

\renewcommand{\arraystretch}{1.15}
\begin{tabular}{lccccccc} % Added one column 'c' for Causal
\toprule
\textsc{Model} & Streaming ? & \textsc{ListOps} & \textsc{Text} & \textsc{Retrieval} & \textsc{Image} & \textsc{Pathfinder} & \textsc{Avg} \\
\midrule
% Global / Non-Causal Baselines
Transformer     & No & 36.37 & 64.27 & 57.46 & 42.44 & 71.40  & 53.66 \\
Reformer        & No & \underline{37.27} & 56.10 & 53.40 & 38.07 & 68.50 & 50.56 \\
BigBird         & No & 36.05 & 64.02 & 59.29 & 40.83 & 74.87  & 54.17 \\
Linear Trans.   & No & 16.13 & \underline{65.90} & 53.09 & 42.34 & 75.30  & 50.46 \\
Performer       & No & 18.01 & 65.40 & 53.82 & 42.77 & 77.05 & 51.18 \\
\midrule
FNet            & No & 35.33 & 65.11 & 59.61 & 38.67 & 77.80  & 54.42 \\
Nystr\"omformer & No & 37.15 & 65.52 & 79.56 & 41.58 & 70.94  & 57.46 \\
Luna-256        & No & 37.25 & 64.57 & 79.29 & \underline{47.38} & 77.72  & 59.37 \\
\midrule
\textbf{S4}     & No & \underline{59.60} & \underline{86.82} & \underline{90.90} & \underline{88.65} & \underline{94.20} & \underline{86.09} \\
% RWKV-v4         & \textbf{Yes} & 55.88 & 86.04 & 88.34 & 70.53 & 58.42 & 72.07 \\
\midrule
% Causal Models (Yours)

% suGRU(binary)   & \textbf{Yes} & \textbf{44.93} & \textbf{64.42} & \textbf{77.54} & \textbf{71.21} & \textbf{84.92} & 68.60 \\ 
suGRU(binary)   & \textbf{Yes} & \textbf{44.93} & \textbf{64.42} & \textbf{77.54} & \textbf{86.37} & \textbf{84.92} & 71.63 \\ 

suGRU(sigmoid)  & \textbf{Yes} & 39.45 & 64.17 & 63.23 & 56.92 & 73.45  & 59.44 \\ 
\bottomrule
\end{tabular}

% \vspace{-\baselineskip}
\end{table*}

\section{Experiments}
 We evaluate the efficacy of Selective-Update RNNs on a comprehensive suite of long-range benchmarks, ranging from the Long Range Arena (LRA) and pixel-level classification to synthetic memory tasks.
\subsection{Long Range Arena}

In the LRA benchmark suite~\cite{tay2020long}, which comprises five tasks with sequence lengths ranging from 1K to 4K steps across multiple modalities and objectives (e.g., similarity matching, structural inference, and visuospatial reasoning), we compare suGRU against 8 Transformer variants and modern linear RNNs (e.g., S4,) as well as more recent methods reported in the literature. As summarized in Table~\ref{tab:LRA}, while models like S4 and LRU~\cite{orvieto2023lru} achieve state-of-the-art accuracy on LRA (~94\% on Pathfinder), they often utilize bidirectional processing or global convolutions for these spatial classification tasks. In contrast, suGRU operates under a strict uni-directional (causal) constraint, processing the sequence strictly token-by-token without access to future context. Despite this handicap, suGRU achieves 84.92\% on Pathfinder, significantly outperforming RWKV-v4 (58.42\%) \cite{peng2023rwkv} and standard causal RNNs which typically fail this task. Ablation results confirm that binary gating decisively outperforms the continuous (sigmoid) alternative.

\subsection{Selective copy}
We evaluate suGRU on the Selective Copy task \cite{gu2024mamba}, a long range synthetic benchmark that requires reading a stream, selecting a small subset of marked symbols, and reproducing them after a long distractor interval ($T = 4096$), making the ideal computation pattern ``sparse write, long carry''. 
% It success depends on stable retention and precise, event triggered updates. 
While, selective updates implement this pattern directly: suGRU can \emph{write} only at marked events and use an \emph{exact carry} ($g_t=0$) through distractors, reducing cumulative state drift and preserving gradient flow.
As shown in Table~\ref{tab:scopy}, selective copy performance is strongly architecture-dependent: Hyena~\cite{poli2023hyena} variants are near chance, S4~\cite{gu2022s4} improves but remains below this setting, while S6~\cite{gu2023mamba} reaches near-perfect accuracy. suGRU is competitive with these strong baselines while remaining strictly streaming: with an S4-style backbone, two layers attain 97.2\% accuracy, and three layers reach 99.5\%. 
This supports our claim that selective updates substantially improve long-sequence credit assignment, and that modest recurrent depth closes the remaining gap.

\begin{table}[t]
\centering
\small
\caption{\textbf{Selective Copy} accuracy across models and configurations. Here, $L$ denotes the number of layers. Reported values are taken from \cite{feng2024minGRU}}
\begin{tabular}{lll}
\toprule
\textbf{Model} & \textbf{Layer} & \textbf{Accuracy} \\
\midrule
H3    & Hyena & 30.1 \\
Mamba & Hyena & 28.4 \\
\midrule
S4    & S4    & 18.3 \\
H3    & S4    & 57.0 \\
Mamba & S4    & 56.4 \\
\midrule
S4    & S6    & 97.0 \\
H3    & S6    & 99.7 \\
Mamba & S6    & 99.8 \\
\midrule
GRU & S4(2$L$) & $16.3 \pm .5$ \\
suGRU & S4(2$L$) & $97.2 \pm .3$ \\
suGRU & S4(3$L$) & $99.5 \pm .5$ \\
\bottomrule
\end{tabular}
\label{tab:scopy}
\end{table}

\subsection{WikiText-103}
WikiText-103 \cite{merity2017pointer} is an established benchmark for language modeling, an important task for large-scale sequence models where tokens are predicted sequentially based on past context. Although RNNs were the model of choice for many years, Transformers are now the dominant model in such applications that contain data that is inherently discrete. We show that sequential RNN models trained via BPTT can still be competitive to Transformers in these settings. 

\begin{figure}[htbp]
  \centering
    \includegraphics[scale = 1.0]{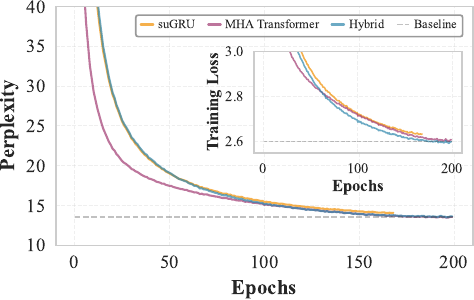}
    \caption{WikiText-103 language modeling learning curves showing training perplexity versus epochs for one-pass suGRU, a Hybrid variant, and a multi-head attention Transformer. The inset reports training loss over the same training run. The dashed horizontal line indicates the baseline transformer performance.}
    \label{fig:wk103}
\end{figure}

As a result, suGRU significantly narrows the performance gap between purely recurrent architectures and attention-based baselines. Under a parameter-matched setting (Table \ref{tab:wikitext103_results}, bottom), suGRU achieves a test perplexity of 21.53, while scaling the model to 100M parameters further reduces perplexity to 18.29. To further enhance performance, we introduce a hybrid architecture that interleaves suGRU layers with classical self-attention. This {\bf Hybrid-suGRU} variant reaches a perplexity of 18.03 with only a marginal increase in parameter count, demonstrating that selective-update recurrence remains highly competitive at the language-modeling scale. As illustrated in Figure~\ref{fig:wk103}, both suGRU and the hybrid variant exhibit optimization stability and convergence rates on par with the Transformer baseline.

\begin{table}[t]
\centering
\small
\caption{\textbf{Language modeling performance on WikiText-103.} Top rows report baseline results are from \cite{qinhgrn2}; bottom rows show our suGRU implementation performance.}

\begin{tabular}{lccc}
\toprule
Model & val PPL ($\downarrow$) & test PPL ($\downarrow$) & Params (M) \\
\midrule
Transformer & 24.40 & 24.78 & 44.65 \\
FLASH       & 25.92 & 26.70 & 42.17 \\
Performer   & 62.50 & 63.16 & 44.65 \\
gMLP        & 28.08 & 29.13 & 47.83 \\
S4          & 38.34 & 39.66 & 45.69 \\
RWKV-4      & 24.31 & 25.07 & 46.23 \\
LRU         & 29.86 & 31.12 & 46.24 \\
Mamba       & 22.58 & 23.19 & 44.39 \\
HGRN2       & 23.10 & 23.73 & 44.66 \\
\midrule
Transformer & 18.32 & 18.44 & \textbf{44.65} \\
% suGRU(k=5)  & 21.47 & 21.53 & 44.86 \\
% suGRU(k=1)  & 19.28 & 19.20 & 44.86 \\
suGRU  & 19.28 & 19.20 & 44.86 \\
suGRU(100M) & 18.29 & 18.29 & 88.98 \\
Hybrid-GRU & \textbf{18.07} & \textbf{18.03} & 48.92 \\
\bottomrule
\end{tabular}
\label{tab:wikitext103_results}
\end{table}

\subsection{Pixel-level classification and analysis}
We conducted more studies on pixel-level classification, from sequential MNIST and permuted variants to sequential CIFAR10.  

\textbf{Performance} On these tasks, suGRU performs strongly under strict sequential training and decoding (Table~\ref{tab:pixel_level_1d_classification}). It matches the best reported accuracy on sMNIST and achieves competitive performance on psMNIST, comparable to Transformer results while retaining constant-memory recurrence. On the sCIFAR task, suGRU substantially outperforms the Transformer baseline and prior recurrent models. 
% Because LSSL \cite{gu2021lssl} is a causal (unidirectional) SSM that models sequences as continuous-time systems, it serves as a more direct and reasonable point of comparison for our architecture than non-causal models \cite{Gu2021S4}. 
We prioritize comparison with LSSL \cite{gu2021lssl} as it represents a more direct, causal SSM baseline that shares our unidirectional, continuous-time modeling constraints.
In addition, suGRU learns faster in these settings, converging in fewer epochs than standard GRU baselines, consistent with the shorter effective credit-assignment paths induced by selective updates. The learning curve and ablations over gate-scheduling variants are reported in the Appendix.

\textbf{Other recurrent baselines.} We evaluate selective updates across several recurrent backbones on psMNIST, including vanilla RNNs, GRU, LSTM, and representative spiking models. Table~\ref{tab:psmnist_app} in Appendix shows consistent improvements over the corresponding non-selective architectures: su-RNN, su-GRU, and su-LSTM substantially outperform their standard counterparts and exceed a range of strong recurrent baselines. Our results show that su-GRU achieves the highest test accuracy among all reported streaming recurrent models. With Selecting Updating applied to SNNs as su-SNN, we establish a new state with 97.33\% accuracy on testset (see Appendix Table \ref{tab:snnpsmnist}), surpassing specialized spiking baselines such as Rhythm-SNNs \cite{yan2025snn2} and Balanced Resonate-and-Fire Neurons \cite{higuchi2024brfsnn}.

\begin{figure}[htbp]
  \centering
    \includegraphics[width=0.45\textwidth]{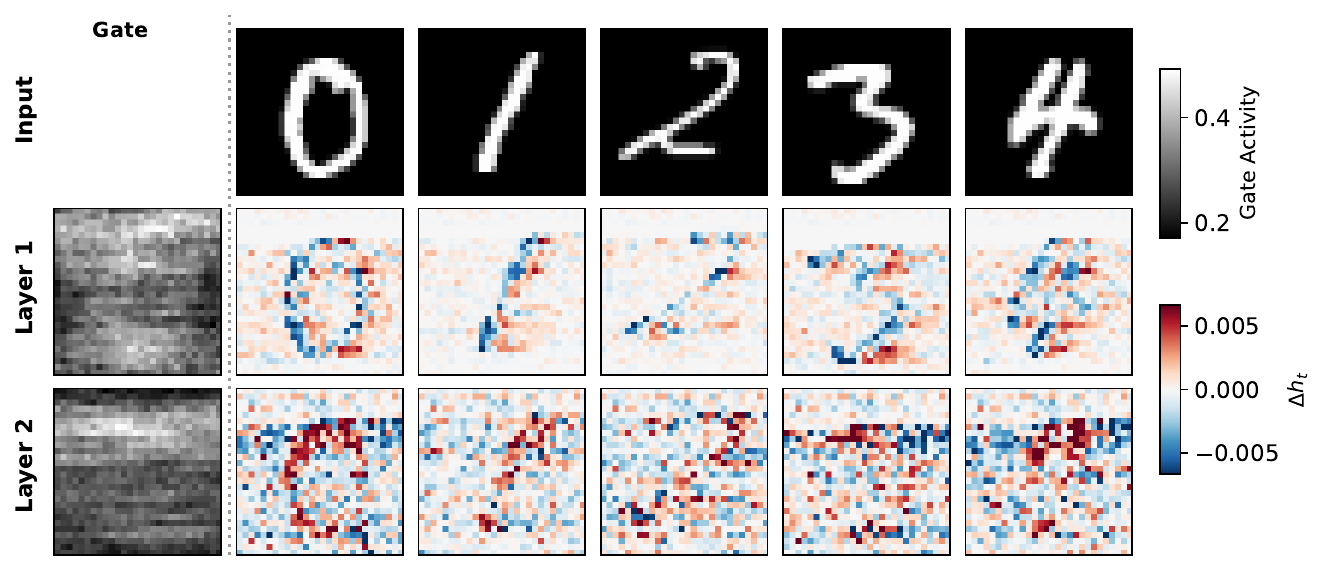}
    
    \caption{
    \textbf{Spatio-temporal dynamics of a two-layer suGRU on sMNIST.}
    Averaged gating activity (left) and hidden state changes $\Delta \bh_t = \bh_t - \bh_{t-1}$ (right) are reshaped to $28 \times 28$ for spatial analysis. The signed increment maps demonstrate that the model performs sparse, event-triggered updates targeted at salient features in both layers, successfully bypassing redundant pixels.
    }
    \vspace{-5pt}
\label{fig:smnist_dynamics}
\end{figure}

\textbf{Feature Visualization} Figure~\ref{fig:smnist_dynamics} contrasts update timing (gates) with update magnitude (state increments) along the input stream. Gates form a stable, layer-dependent schedule with a moderate opening rate, largely consistent across digits. Increments $\Delta h_t$ are concentrated at edge transitions and high-contrast changes, while background regions produce near-carry behavior. The second layer exhibits more distributed updates than the first, consistent with deeper recurrence integrating local evidence into higher-level state changes.

% \paragraph{Ablation} comparing the binary gate with sigmoid gate, same network with sigmoid gate only got 73\% on pathfinder task instead of 84.92\% with binary gate. 

\begin{table}[t]
\centering
\small
\caption{\textbf{Pixel-level 1-D image classification.} Comparison against reported test accuracies from prior works \cite{gu2022s4}.}
\label{tab:pixel_level_1d_classification}
\begin{tabular}{lccc}
\toprule
 & sMNIST & psMNIST & sCIFAR \\
\midrule
Transformer     & 98.9  & 97.9  & 62.2  \\
\midrule
LSTM            & 98.9  & 95.11 & 63.01 \\
r-LSTM          & 98.4  & 95.2  & 72.2 \\
UR-LSTM         & 99.28 & 96.96 & 71.00 \\
UR-GRU          & 99.27 & 96.51 & 74.4  \\
HiPPO-RNN       & 98.9  & 98.3  & 61.1  \\
LMU-FFT         & --    & 98.49 & --    \\
LipschitzRNN    & 99.4  & 96.3  & 64.2  \\
\midrule
TCN             & 99.0  & 97.2  & --    \\
TrellisNet      & 99.20 & 98.13 & 73.42 \\
CKConv          & 99.32 & 98.54 & 63.74 \\
LSSL            & 99.53 & 98.76 & 84.65 \\
\midrule
\textbf{suGRU}     & \textbf{99.53}    & \textbf{98.86} & \textbf{89.21} \\
\bottomrule
\end{tabular}
\end{table}

 \vspace{-15pt}
\section{Related work}
% 1) depth residual connection
% 2) SA and BPTT
% 3) other temporal residual connection, most of them are additive
% 4) selective update in the memory and bio rhythmic used to activate the brain area 

Residual connections address the optimization difficulties of very deep networks by making identity propagation the default and learning only deviations, which creates short gradient routes through depth and mitigates degradation as layers increase \cite{He2016ResNet,Veit2016EnsembleResNet}. Attention offers an analogous benefit in sequence models by shortening token-to-token credit assignment with direct interactions rather than long temporal chains \cite{Vaswani2017Transformer}. Our work follows the same principle along the temporal axis of recurrence, aiming to reduce the number of nontrivial transformations that gradients must traverse while preserving a strictly streaming recurrent computation.

Several recurrent research lines incorporate temporal residual structure, but they differ in how identity is introduced and in what is being optimized. Traditional additive temporal residual and highway-style RNNs \cite{Srivastava2015Highway} increase trainability by introducing skip pathways across depth or time while still applying a recurrent transform at every step, and stochastic carry regularizers such as Zoneout \cite{Krueger2017Zoneout} encourage partial identity behavior but do not explicitly control when computation is executed.
% The second line explicitly learns to skip updates, including Adaptive Computation Time \cite{graves2016adaptive} and Skip RNN \cite{campos2018skip}, as well as multi-rate schedules such as Clockwork RNN \cite{Koutnik2014Clockwork}, Phased LSTM \cite{Neil2016PhasedLSTM}, and hierarchical multiscale RNNs \cite{Chung2016HMRNN}. 
As noted earlier, the second line explicitly learns to skip updates, including Adaptive Computation Time \cite{graves2016adaptive} and Skip RNN \cite{campos2018skip}, as well as multi-rate schedules such as Clockwork RNN \cite{Koutnik2014Clockwork}, Phased LSTM \cite{Neil2016PhasedLSTM}, and hierarchical multiscale RNNs \cite{Chung2016HMRNN}.
These methods typically gate whole-state updates or introduce halting and regularization objectives to balance accuracy and compute, and they often rely on sensitive training heuristics or fixed scheduling assumptions. In contrast, our selective updates introduce per-unit identity-through-time carries, preserve the base recurrent transform, and admit a direct path-length analysis in terms of updates.

\section{Discussion and Conclusion}

We introduced Selective-Update RNNs (suRNNs) to mitigate memory decay by decoupling update frequency from sequence length via binary gating. This mechanism is inspired by biological models of working memory, where frontostriatal circuits explicitly learn when to update internal representations versus maintain them \cite{Frank2001CABN,ChathamFrankBadre2014Neuron,BhandariBadre2018GatingPolicies}. 

A key contribution of this work is demonstrating strong performance under strict \textit{uni-directional} (causal) constraints. While non-causal State Space Models (e.g., S4, Mamba) excel on the Long Range Arena by leveraging non-causal convolutions or bi-directional views, suGRU operates as a strictly uni-directional streaming model. Our 84.92\% accuracy on \textsc{Pathfinder} shows that strictly causal RNNs can capture complex long-range spatial dependencies if the vanishing gradient problem is structurally mitigated. By implementing this theoretical improvement via a cuDNN fusion, we demonstrated that biologically plausible sparsity can be implemented without sacrificing hardware efficiency.

Our results suggest three directions for future research. First, we relied on standard GRU backbones to isolate the gating effect; more expressive recurrent parameterizations or context-aware gate generators could further raise the performance ceiling. Second, training still relies on BPTT, which remains a bottleneck for extreme lengths; future work should explore event-driven backpropagation or sparse checkpointing to better exploit temporal sparsity during optimization. Third, extending selective updates to bi-directional architectures could enhance performance for non-causal tasks with global context. Finally, the induced subnetwork structure suggests a natural connection to continual learning \cite{kang2022clsub}, where separating updates across contexts could reduce interference between tasks. Ultimately, taking inspiration from the biological principle of selective maintenance with the hardware efficiency of one-pass recurrence, suRNNs re-establish the viability of strict streaming architectures for long-context learning.

\newpage

% \section*{Acknowledgments}
% This was was supported in part by......

% \section*{Impact Statement}

% This paper presents work whose goal is to advance the field of Deep Learning. There are many potential societal consequences of our work, none which we feel must be specifically highlighted here.

\nocite{langley00}

\bibliography{references}
\bibliographystyle{icml2026}

%%%%%%%%%%%%%%%%%%%%%%%%%%%%%%%%%%%%%%%%%%%%%%%%%%%%%%%%%%%%%%%%%%%%%%%%%%%%%%%
%%%%%%%%%%%%%%%%%%%%%%%%%%%%%%%%%%%%%%%%%%%%%%%%%%%%%%%%%%%%%%%%%%%%%%%%%%%%%%%
% APPENDIX
%%%%%%%%%%%%%%%%%%%%%%%%%%%%%%%%%%%%%%%%%%%%%%%%%%%%%%%%%%%%%%%%%%%%%%%%%%%%%%%
%%%%%%%%%%%%%%%%%%%%%%%%%%%%%%%%%%%%%%%%%%%%%%%%%%%%%%%%%%%%%%%%%%%%%%%%%%%%%%%
\newpage
\appendix
\setcounter{figure}{0}
\renewcommand{\thefigure}{A\arabic{figure}}

\setcounter{table}{0}
\renewcommand{\thetable}{A\arabic{table}}

\onecolumn

% You can have as much text here as you want. The main body must be at most $8$
% pages long. For the final version, one more page can be added. If you want, you
% can use an appendix like this one.

% The $\mathtt{\backslash onecolumn}$ command above can be kept in place if you
% prefer a one-column appendix, or can be removed if you prefer a two-column
% appendix.  Apart from this possible change, the style (font size, spacing,
% margins, page numbering, etc.) should be kept the same as the main body.
%%%%%%%%%%%%%%%%%%%%%%%%%%%%%%%%%%%%%%%%%%%%%%%%%%%%%%%%%%%%%%%%%%%%%%%%%%%%%%%
%%%%%%%%%%%%%%%%%%%%%%%%%%%%%%%%%%%%%%%%%%%%%%%%%%%%%%%%%%%%%%%%%%%%%%%%%%%%%%%
\section{su-RNNs}
\begin{figure*}[t]
    \centering
    \includegraphics[width=1\linewidth]{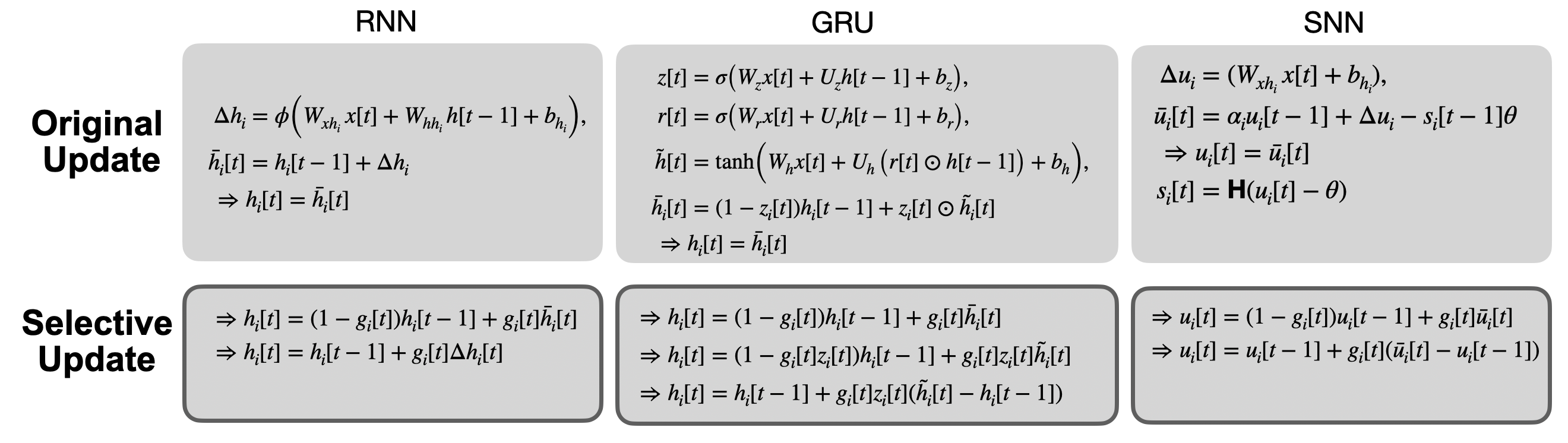}
    \caption{\textbf{Selective-update across recurrent families.} Top: the update equations for vanilla RNN, GRU, and a spiking neuron model (SNN). Bottom: the same modules rewritten in selective-update form by inserting a per-unit gate $g_i[t]$. In GRU, it composes with the native continuous gate $z_i[t]$, and in SNN it acts on membrane integration. This exposes a unified timing control across architectures while leaving the underlying transforms unchanged.}
    \label{fig:rnn_summary}
\end{figure*}

Figure \ref{fig:rnn_summary} demonstrates the universality of the selective-update mechanism across three distinct recurrent families: the vanilla RNN, the Gated Recurrent Unit (GRU), and the Spiking Neural Network (SNN). In each architecture, the binary gate $g_i[t]$ acts as a residual switch that arbitrates between the standard state evolution and an exact identity carry. For the RNN, it gates the non-linear residual $\Delta h_t$; for the GRU, it effectively masks the native continuous update gate $z_t$; and for the SNN, it regulates the integration of the membrane potential. This unified formulation illustrates that selective updates serve as a backbone-agnostic control layer, effectively decoupling the timing of updates from the specific dynamical laws of the underlying neuron.

\section{Efficient Implementation of suGRU}
\label{Apx:effsuGRU}
\begin{figure}

% \begin{wrapfigure}{r}{0.5\textwidth}
  \centering
  \includegraphics[width=0.48\textwidth]{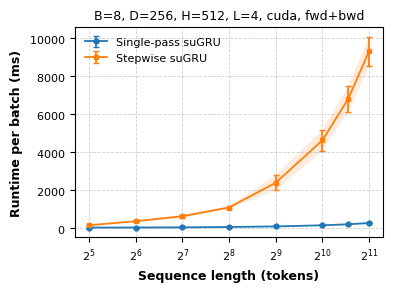}
  \caption{Runtime per batch (ms) versus sequence length for two suGRU implementations. \textbf{Single-pass suGRU} runs the entire sequence in one cuDNN GRU call using the skip-drive input, yielding near-flat scaling and substantially lower wall-clock time at long sequences. \textbf{Stepwise suGRU} applies per-step control flow and incurs rapidly growing cost. Error bars indicate run-to-run variation.}
  % \vspace{-.05cm}
  \label{fig:sugru-speed}
% \end{wrapfigure}
\end{figure}
BPTT is often viewed as too slow to scale across long sequences and deep networks. Building on the selective-update concept, we implement \emph{suGRU} so that it executes in a \emph{single} cuDNN GRU pass while realizing a temporal-residual update/carry at each step. We supply an external binary gate $g_t\in\{0,1\}^H$ (1\,=\,update, 0\,=\,carry) and concatenate the \emph{skip drive} $(1-g_t)$ as $H$ additional input channels. Inside the fused GRU we \emph{hard-wire only the $z$-head} on these extra channels to a fixed diagonal $+C\,I_H$ (the $r/n$ heads are wired to zero; all other weights remain trainable. More details are explained in Appendix \ref{Apx:suGRU}). This equation represents a $z$-preactivation shift:
$$
\mathbf{z}_t^{\text{pred}} = \mathbf{z}_t^{\text{org}} + C(\mathbf1 - \bg_t),
$$
Here, $\mathbf{z}_t^{org}$ denotes the $z$-preactivation in the original GRU implementation.
so that $g_t=0$ yields a step \emph{identical} to a standard GRU (exact update), while $g_t=1$ saturates the $z$-gate toward $1$ and behaves as a carry in the residual form $h_t \approx h_{t-1}$. Gates are produced once per sequence by lightweight, length-agnostic generators (e.g., B-splines with compact support, time-warped sines, or spectral masks) trained with a straight-through estimator. Because no custom control flow is introduced, the entire sequence is executed in a single cuDNN kernel with GRU throughput. The only runtime overhead comes from the $+H$ gate channels. The asymptotic complexity remains $O\!\big(T(HD+H^2)\big)$ without adding any extra parameters to the GRU component. As a result, Fig.~\ref{fig:sugru-speed} shows a significant speedup compared to vanilla step-wise updates. The code is avaiable at \url{https://anonymous.4open.science/r/suGRU-EB5C}.

The same implementation supports \emph{group gating} (one gate shared by a block of units), an \emph{always-on subset} (guaranteeing a live gradient path), and \emph{state expansion} via grouped projections, without breaking cuDNN fusion. In effect, suGRU preserves GRU-level efficiency while converting optimization depth from sequence length to the \emph{number of updates}, enabling scalable BPTT on long sequences at reasonable speed.

\subsection{Validating the cuDNN-fused suGRU via the \texttt{dense\_z} map}
\label{Apx:suGRU}
\paragraph{Setup (augment the input with a skip-drive).}
We form an augmented input
\[
\mathbf{\hat{x}}_t \;=\; \begin{bmatrix} \mathbf{x}_t \\ \mathbf{I}_{H} - \mathbf{g}_t \end{bmatrix} \in \mathbb{R}^{D+H},
\]
where $g_t \in \{0,1\}^H$ (1 = update, 0 = carry). The GRU preactivations for the three heads (reset $r$, update $z$, candidate $n$) are
\[
\begin{bmatrix}
\mathbf{r}_t^{\mathrm{pre}} \\[2pt]
\mathbf{z}_t^{\mathrm{pre}} \\[2pt]
\mathbf{n}_t^{\mathrm{pre}}
\end{bmatrix}
\;=\;
\mathbf{W}_{ih}\,\mathbf{\hat{x}}_t \;+\; \mathbf{W}_{hh}\,\mathbf{h}_{t-1} \;+\; \mathbf{b}_{ih} + \mathbf{b}_{hh},
\qquad
\mathbf{W}_{ih}\in\mathbb{R}^{3H\times(D+H)},\;\; \mathbf{W}_{hh}\in\mathbb{R}^{3H\times H}.
\]

Partition $\mathbf{W}_{ih}$ by rows (heads) and columns (real inputs vs.\ gate channels):
\[
W_{ih} \;=\;
\begin{bmatrix}
\mathbf{W}^{(r)}_{x} & \mathbf{W}^{(r)}_{g}\\[2pt]
\mathbf{W}^{(z)}_{x} & \mathbf{W}^{(z)}_{g}\\[2pt]
\mathbf{W}^{(n)}_{x} & \mathbf{W}^{(n)}_{g}
\end{bmatrix},
\quad
\mathbf{W}^{(\cdot)}_{x}\in\mathbb{R}^{H\times D},\;\;
\mathbf{W}^{(\cdot)}_{g}\in\mathbb{R}^{H\times H}.
\]

\paragraph{Forced wiring on the gate channels.}
We enforce
\[
\mathbf{W}^{(r)}_{g} = 0,\qquad
\mathbf{W}^{(n)}_{g} = 0,\qquad
\boxed{\,\mathbf{W}^{(z)}_{g} = C\,\mathbf{I}_H\,},\quad C<0,
\]
while all other blocks remain trainable. Forced wiring is applied following each training update.

\paragraph{What \texttt{dense\_z} computes under the wiring.}
With the above structure, the $z$-preactivation satisfies
\[
\begin{aligned}
\mathbf{z}_t^{\mathrm{pre}}
&= \mathbf{W}^{(z)}_{x}\,\mathbf{x}_t \;+\; \mathbf{W}^{(z)}_{g}\,(\mathbf{I}_H-\mathbf{g}_t) \;+\; \mathbf{W}^{(z)}_{h}\,\mathbf{h}_{t-1} \;+\; \mathbf{b}^{(z)}\\
&= \underbrace{\big(\mathbf{W}^{(z)}_{x}\mathbf{x}_t + \mathbf{W}^{(z)}_{h}\mathbf{h}_{t-1} + \mathbf{b}^{(z)}\big)}_{=:z_t^{org}}
\;+\; \underbrace{C\,\mathbf{I}_H}_{W^{(z)}_{g}}\,(\mathbf{I}_H-\mathbf{g}_t)\\[2pt]
&= \boxed{\,\mathbf{z}_t^{org} + C(\mathbf{I}_H-\mathbf{g}_t)\,},
\end{aligned}
\]
where $\mathbf{z}_t^{org}$ represents the preactivation of the original GRU implementation.
Thus, the update gate is effectively modified as 
$$
\mathbf{z}'_t \;=\; \sigma\!\big(\mathbf{z}_t^{org} + C(\mathbf{I}_H-\mathbf{g}_t)\big),
$$
While the $r$ and $n$ heads remain unchanged (with their gate-channel blocks set to zero), the proposal $\tilde h_t$ is computed in the same manner as a standard GRU. 

\paragraph{suGRU update and comparison with a hard temporal residual.}
The GRU state update in residual form is
\[
\mathbf{h}_t \;=\; \mathbf{h}_{t-1} \;+\; \mathbf{z}'_t\,\odot\,\big(\tilde h_t - h_{t-1}\big).
\]
% A \emph{hard} selective residual would use
% \[
% h_t^{\mathrm{hard}} \;=\; h_{t-1} \;+\; g_t \odot (\tilde h_t - h_{t-1}),
% \quad\Rightarrow\quad
% w^{\mathrm{hard}}_t = g_t.
% \]
Using $\mathbf{z}'_t=\sigma(\mathbf{z}_t^{org} + C(\mathbf{I}_H-\mathbf{g}_t))$, the gate signal is obtained:
\[
\boxed{\;
\mathbf{z}'_t
% = 1 - \sigma\!\big(a_t + C(1-g_t)\big)
= \mathbf{g}_t\sigma(\mathbf{z}_t^{org}) \;+\;
(1-\mathbf{g}_t)\;\sigma(\mathbf{z}_t^{org} + C)}, \; \text{where} \;{C < 0}\;.
\]
Hence:
\begin{itemize}
\item \textbf{Update steps} ($g_t=1$): $\mathbf{z}'_t = \sigma(\mathbf{z}_t^{org})$ — the step is \emph{identical} to a vanilla GRU update where .
\item \textbf{Carry steps} ($g_t=0$): $\mathbf{z}'_t = \sigma(\mathbf{z}_t^{org}+C\mathbf{I}_H) \xrightarrow{} 0$ — a nonnegative, exponentially small deviation from a perfect carry. 
\end{itemize}

\paragraph{Retention over long carry runs and choice of $C$.}
Let $a_t := -z_t^{\mathrm{org}}$ denote the carry margin, so that $1-z'_t=\sigma(a_t-C)$. Over $L$ consecutive carry steps, the retained mass satisfies
\[
\prod_{k=1}^{L} (1-z'_{t-k+1})
\;=\;
\prod_{k=1}^{L} \sigma(a_{t-k+1}-C)
\;\ge\;
\sigma(a_{\min}-C)^L,
\]
whenever $a_{t}\ge a_{\min}$ holds on carry steps. To ensure that the retained mass after $L$ carries is at least $1-\rho$, it suffices to choose $C$ such that
\[
\sigma(a_{\min}-C)^L \;\ge\; 1-\rho
\quad\Longleftrightarrow\quad
C \;\le\; a_{\min} - \mathrm{logit}\!\big((1-\rho)^{1/L}\big),
\qquad
\mathrm{logit}(p):=\log\frac{p}{1-p}.
\]
This design aims to bring $z'_t$ closer to $0$ in carry steps, thereby facilitating the efficient transmission of $h_{t-1}$ over long carry sequences. The significance of normalization is underscored in this context: by stabilizing the distribution of $z_t^{\mathrm{org}}$, it ensures a steady margin $a_t=-z_t^{\mathrm{org}}$, thus enabling the realization of a consistent retention guarantee with a constant $C$.

\section{Clarification of Proposition 1 and the constant $C_i$}
\label{app:Ci}

In this section, we clarify the row-wise gradient argument underlying Proposition~1. First, recall the selective-update recurrence
\begin{equation}
h_t
=
h_{t-1}
+
D_t\bigl(f_\theta(x_t,h_{t-1}) - h_{t-1}\bigr),
\label{eq:sur_update_appendix}
\end{equation}
where $D_t=\mathrm{diag}(g_t)\in\{0,1\}^{H\times H}$. Define
\begin{equation}
J_t
:=
\frac{\partial h_t}{\partial h_{t-1}}
=
I + D_t\bigl(J_t^{(f)}-I\bigr),
\qquad
J_t^{(f)}
:=
\frac{\partial f_\theta(x_t,h_{t-1})}{\partial h_{t-1}}.
\label{eq:Jt_decomp_appendix}
\end{equation}
Hence, for any $s<t$,
\begin{equation}
M_{s,t}
:=
\frac{\partial h_t}{\partial h_s}
=
J_tJ_{t-1}\cdots J_{s+1}.
\label{eq:Ms_t_def}
\end{equation}

For neuron $i$, we study the sensitivity of the $i$-th hidden coordinate at time $t$ to the hidden state at time $s$, namely the $i$-th row of $M_{s,t}$:
\begin{equation}
r_{s,t,i}
:=
e_i^\top M_{s,t}
=
e_i^\top \frac{\partial h_t}{\partial h_s}
\in \mathbb{R}^{1\times H},
\label{eq:rsti_def}
\end{equation}
where $e_i\in\mathbb{R}^H$ is the $i$-th standard basis vector.

Using $M_{s,t}=J_tM_{s,t-1}$, we obtain
\begin{equation}
r_{s,t,i}
=
e_i^\top J_t M_{s,t-1}.
\label{eq:rsti_recursion_start}
\end{equation}
The key observation is that, because $D_t$ is diagonal, the $i$-th row of $J_t$ depends only on the scalar gate $g_{t,i}$:
\begin{equation}
e_i^\top J_t
=
e_i^\top\!\left(I + D_t(J_t^{(f)}-I)\right)
=
\begin{cases}
e_i^\top, & g_{t,i}=0,\\[2mm]
e_i^\top J_t^{(f)}, & g_{t,i}=1.
\end{cases}
\label{eq:row_case_split}
\end{equation}
Substituting \eqref{eq:row_case_split} into \eqref{eq:rsti_recursion_start} yields
\begin{equation}
r_{s,t,i}
=
\begin{cases}
r_{s,t-1,i}, & g_{t,i}=0,\\[2mm]
e_i^\top J_t^{(f)}M_{s,t-1}, & g_{t,i}=1.
\end{cases}
\label{eq:correct_row_recursion}
\end{equation}

Equation~\eqref{eq:correct_row_recursion} precisely captures the role of selective updates in the row-wise gradient dynamics. When $g_{t,i}=0$, the $i$-th row is propagated through the identity row of $J_t$, so time $t$ does not introduce a new non-identity Jacobian factor into the row-$i$ sensitivity. When $g_{t,i}=1$, the update Jacobian $J_t^{(f)}$ enters.

Let
\begin{equation}
U(s,t)
:=
\{\tau\in\mathbb{Z}: s<\tau\le t\},
\qquad
U_i^{\mathrm{on}}(s,t)
:=
\{\tau\in U(s,t): g_{\tau,i}=1\}.
\label{eq:Ui_on_def}
\end{equation}
Only times in $U_i^{\mathrm{on}}(s,t)$ contribute new non-identity factors to the row-$i$ evolution. Therefore, if
\begin{equation}
\|J_\tau^{(f)}\|\le \rho
\qquad
\text{for all }\tau\in U(s,t),
\label{eq:Jf_bound}
\end{equation}
for any induced matrix norm $\|\cdot\|$, then
\begin{equation}
\|r_{s,t,i}\|
\le
C_i(s,t)\,\rho^{|U_i^{\mathrm{on}}(s,t)|},
\label{eq:row_bound_general}
\end{equation}
where $C_i(s,t)$ absorbs cross-coordinate coupling accumulated after update times. In particular, if those carry-only propagations are uniformly bounded, then $C_i(s,t)\le C_i$ for some constant $C_i$ independent of $t-s$, and thus
\begin{equation}
\left\|
e_i^\top \frac{\partial h_t}{\partial h_s}
\right\|
\le
C_i\,\rho^{|U_i^{\mathrm{on}}(s,t)|}.
\label{eq:row_bound_final}
\end{equation}

Equation~\eqref{eq:row_bound_final} justifies the statement in Proposition~1: the exponent governing amplification or attenuation along neuron $i$ depends on the number of times that neuron is updated, $|U_i^{\mathrm{on}}(s,t)|$, rather than on the elapsed number of steps $t-s$. In a conventional dense RNN, every neuron updates at every step, so $|U_i^{\mathrm{on}}(s,t)|=t-s$, recovering the usual bound.

\paragraph{Effective depth and update rate.}
The average per-neuron update rate over a sequence of length $T$ is
\begin{equation}
p
:=
\frac{1}{HT}\sum_{i=1}^H |U_i^{\mathrm{on}}(0,T)|
\in [0,1].
\label{eq:update_rate_corrected}
\end{equation}
Hence the expected number of updates per neuron is $pT$. Substituting this into \eqref{eq:row_bound_final} gives the heuristic scaling
\begin{equation}
\left\|
\frac{\partial h_T}{\partial h_0}
\right\|
\lesssim
C\,\rho^{\,pT},
\label{eq:heuristic_scaling}
\end{equation}
for an appropriate constant $C$, making explicit that reducing the update rate reduces the effective multiplicative depth of the recurrent computation.

We emphasize that this argument is row-wise: identity carries remove non-identity Jacobian factors from the sensitivity path of a given neuron when that neuron does not update. The analysis does not eliminate cross-coordinate coupling altogether; rather, such coupling can only enter through update times and is absorbed into the constant $C_i$ (or $C_i(s,t)$).

\newpage

\section{Stepwise Result}
We first study selective updates in the standard stepwise RNN implementation on well-known synthetic and sequential benchmarks, to isolate the source of the performance gains and to better understand the underlying mechanism.

\subsubsection{Copying-Memory task }
We revisit the well-known Copying-Memory task \cite{le2015simple}, where the model observes a short prefix, then $T$ blanks and a delimiter, and must reproduce the prefix after the delimiter, stressing long-range dependency. For sequences with delay $T$, we track $\|\partial \ell/\partial h_\tau\|$ as a function of temporal distance for both RNN and GRU,  and their selective-update variants. With selective updates, in Fig. \ref{fig:copy_task} a), gradient traces remain bounded and nearly parallel as $T$ increases, consistent with the Jacobian structure $J_t = I + D_t(J_f^t - I)$ and Proposition~1, which predicts effective multiplicative depth governed by updates taken rather than input length. On Copying-Memory with $T=5000$, the gated models converge faster and reach lower loss, indicating time-sparse credit concentrated on update steps and empirically validating the theoretical mechanism.

\begin{figure}[htbp]
  \centering
  \begin{minipage}{0.75\textwidth}
    \includegraphics[width=\textwidth]{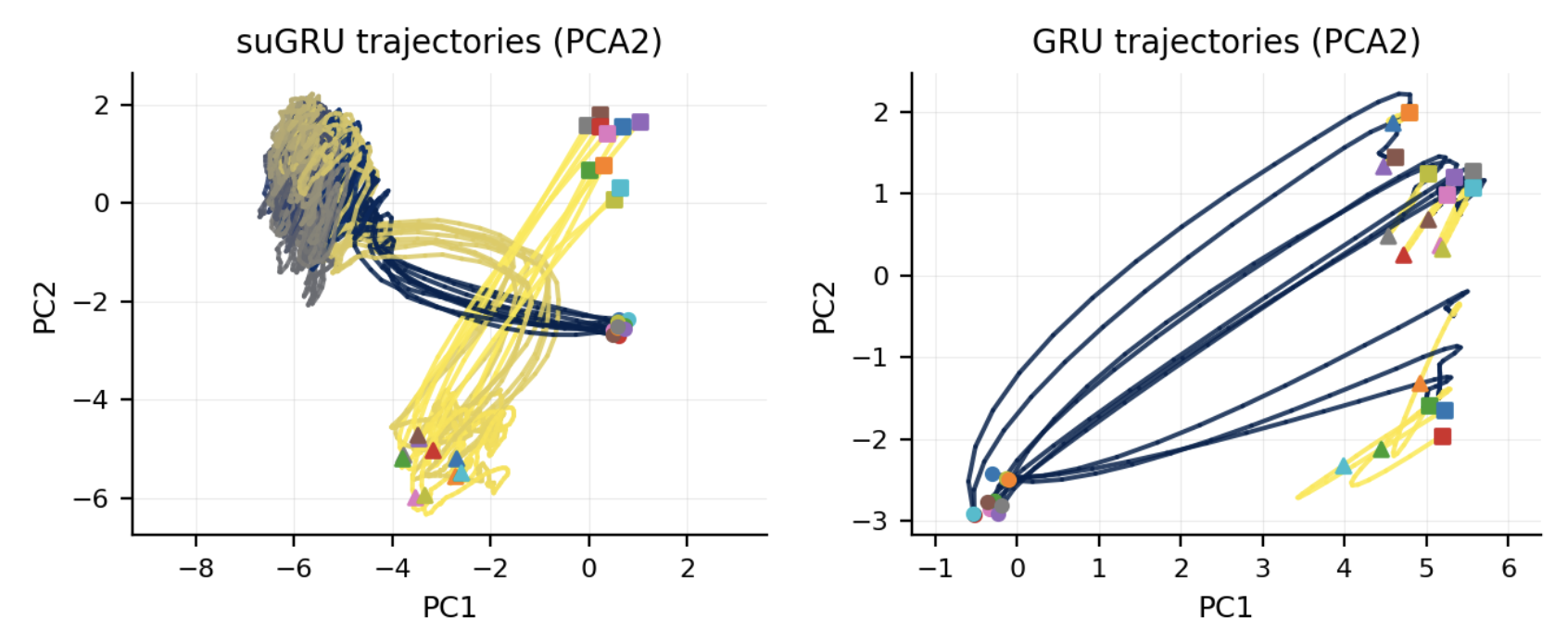}
  \end{minipage}%
  \caption{\textbf{PCA projection of hidden-state trajectories on Copy-Memory ($T{=}2000$).} Starting from the initial hidden state (circle), we follow the recurrent dynamics in the 2D plane; the delimiter timestep is marked by a triangle and the final timestep by a square. Each curve corresponds to one sequence (10 examples in total). Line color encodes normalized time, from early (dark) to late (bright). suGRU (left) exhibits more structured evolution with a clearer separation between delay and recall regimes compared to GRU (right).}
        \label{fig:copy_task_traj}
  % \end{minipage}
\end{figure}

% In Fig~\ref{fig:copy_task_traj}, we visualize the internal dynamics by projecting hidden states $h_t$ onto the first two principal components computed from the collected state trajectories. For $T = 2000$, the GRU trajectories shown a pronounced drift during the delay phase, spanning a large region in the PCA plane before entering recall. In contrast, suGRU trajectories remain more localized during delay and display a sharper transition at the delimiter into recall-associated regions. These qualitative differences indicate that suGRU organizes its state space with stronger phase separation between “maintenance” and “readout” dynamics under long delays.

% In the copy task, the delay region is filled with zeros during training, so a GRU can implicitly use this low-entropy segment as a timing cue. We stress-test this by replacing the zero delay with random noise at test time. The GRU collapses to near chance (about 60\% to 10\%), while the selective-update model degrades more gracefully (about 85.31\% to 48.56\%). This behavior is consistent with selective updates carrying much of the state unchanged through the delay, making retention less sensitive to the un-selected input statistics.

Figure~\ref{fig:copy_task_traj} projects hidden-state trajectories onto the first two principal components. For long delays ($T=2000$), GRU states exhibit substantial drift throughout the delay phase, covering a broad region of the PCA plane before recall, whereas suGRU remains tightly clustered during delay and transitions sharply at the delimiter into recall-specific regions, suggesting clearer separation between “maintenance” and “readout” dynamics. Because the copy-task delay is filled with zeros during training, a GRU may exploit this low-entropy segment as an implicit timing cue; 
\begin{wrapfigure}{r}{0.48\textwidth}
  \centering
  \vspace{-0.5\baselineskip}
  \includegraphics[width=0.46\textwidth]{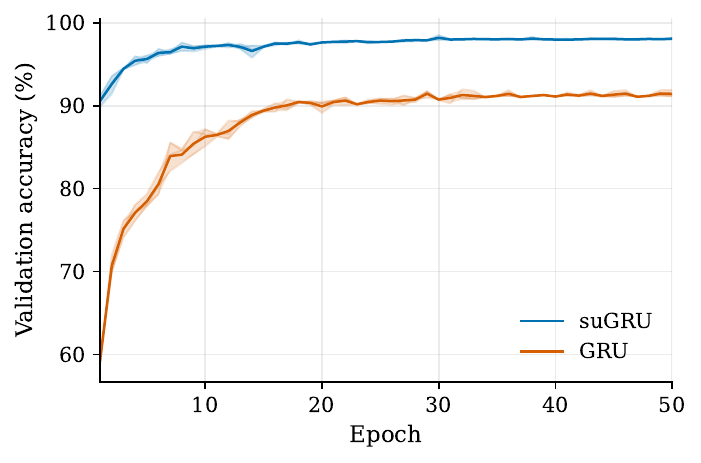}
  \caption{\textbf{Learning curves on psMNIST.} Validation accuracy vs.~epoch for GRU and suGRU under the same training setting; suGRU converges faster and attains higher accuracy.}
  \label{fig:psmnsit_lc}
  \vspace{-2.\baselineskip}
\end{wrapfigure}
replacing the zero delay with random noise at test time confirms this sensitivity: GRU accuracy drops from 27.81\% to $\sim$10\%, while the selective-update model degrades more gradually (85.31\% to 48.56\%), consistent with selective updates preserving state through the delay and reducing dependence on delay-input statistics.

\subsubsection{psMNIST}

\begin{table}[t]
\centering
\caption{Validation and test set accuracy for psMNIST.}
\label{tab:psmnist_app}
% -------- Left: non-spiking --------
\begin{subtable}[htbp]{0.48\textwidth}
\centering
\begin{tabular}{lcc}
\hline
Model        & Validation & Test \\
\hline
FF-baseline  & 92.37 & 92.65 \\
RNN-orth     & 88.70 & 89.26 \\
RNN-id       & 85.98 & 86.13 \\
LSTM         & 90.01 & 89.86 \\
LSTM-chrono  & 88.10 & 88.43 \\
GRU          & 92.16 & 92.39 \\
JANET        & 92.50 & 91.94 \\
SRU          & 92.79 & 92.49 \\
GORU         & 86.90 & 87.00 \\
NRU          & 95.46 & 95.38 \\
Phased LSTM  & 88.76 & 89.61 \\
LMU          & 96.97 & 97.15 \\
\hline
su-RNN       & 97.80 & \textbf{97.76} \\
su-GRU       & 97.93 & \textbf{98.21} \\
su-LSTM      & 97.80 & 97.64             \\
\hline
su-GRU(fixed random Rhy) & \textbf{97.44} & \textbf{97.73} \\
su-GRU(every 3 steps) & 10 & 10. \\
su-GRU(fixed random ) & 10. & 10. \\
su-GRU(x$>$0) & nan & nan \\
su-GRU(learnable $N*T$) & 13.86 & 13.25 \\

\hline
\end{tabular}
\subcaption{Non-spiking neural networks. Ablation study on the basic gate design. 
\textit{Fixed random Rhythm} uses fixed, randomly initialized parameters for gate generation. 
\textit{Fixed random} samples gates from a uniform distribution and then binarizes them via a threshold; 
\textit{$x>0$ gate} defines the gate directly from the raw input: whenever a pixel value is positive, the gate is open and the network state is updated.}
\end{subtable}
\hfill
% -------- Right: spiking --------
\begin{subtable}[t]{0.48\textwidth}
\centering
\begin{tabular}{lcc}
\hline
Model        & Validation & Test \\
\hline
GLIF         & --         & \textbf{90.47} \\
SRNN         & --         & \textbf{94.3}  \\
brf-SNN      & --         & \textbf{95.2}  \\
ASRC-SNN     & --         & \textbf{96.62}  \\
Rhythm-SNNs  & --         & \textbf{96.73} \\
\hline
su-SNN       & \textbf{97.21} & \textbf{97.33}$\pm$ 0.32 \\
\hline
\end{tabular}
\subcaption{Spiking neural networks}
\end{subtable}
\label{tab:snnpsmnist}
\end{table}

Permuted sequential MNIST (psMNIST) is one of the widely used long-range dependency benchmark for recurrent models, where each digit is streamed as a length-784 sequence under a fixed random permutation and the classifier must integrate information across the full horizon. Table 1(a) shows that selective-update recurrence consistently improves classical RNNs and gated variants, with su-RNN and su-GRU reaching the best non-spiking accuracy among the compared baselines and exceeding strong recurrent and state-space  models such as LMU. The ablations isolate the source of the gain: a fixed random rhythmic schedule remains competitive but degrades relative to learned rhythms, while naive gates either collapse to chance or fail to optimize, indicating that correct update timing is essential for stable training. Table 1(b) further shows that the same principle transfers to spiking settings, where su-SNN improves over previous spiking sota performance, suggesting that selective updates provide a general mechanism for long-horizon credit assignment beyond a particular neuron model.

% \paragraph{Feature and gate visualization}, including learning curves, gate sparsity, hidden states and grad norm. 

\begin{figure}[htbp]
  \centering
    \includegraphics[width=\textwidth]{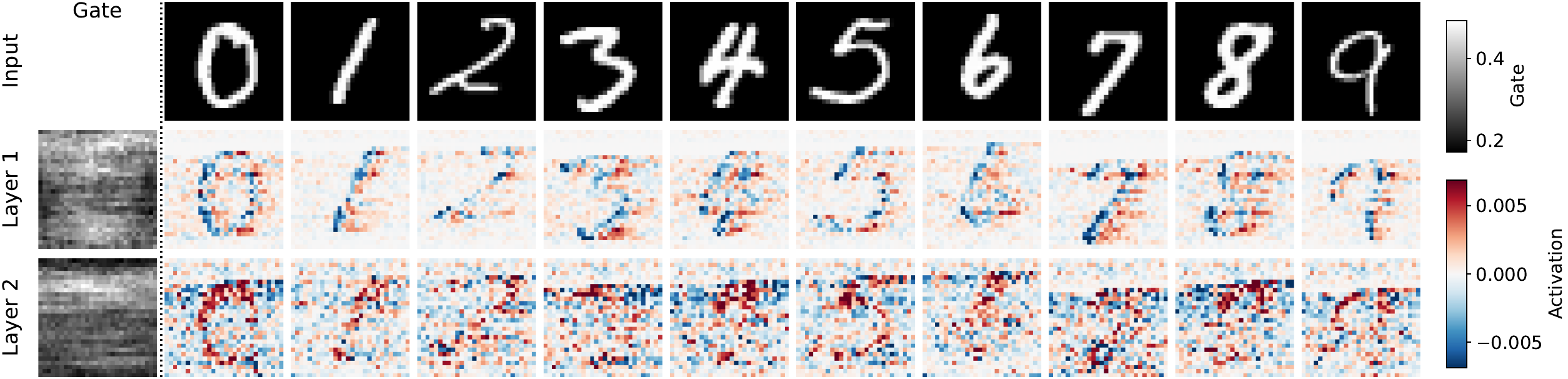}
    \caption{\textbf{Selective-update dynamics on sMNIST.} We visualize a two-layer suGRU with 256 hidden units per layer on the sequential MNIST pixel stream. For readability, the $784$-step input and the corresponding per-timestep quantities are reshaped to $28\times 28$. The left column shows the learned gate activity for each layer (gate value averaged over units), while the remaining columns show, for each digit, the input (top) and the signed hidden-state increment $\Delta h_t = h_t - h_{t-1}$ in Layer~1 and Layer~2 (bottom), highlighting when the model actually modifies its recurrent state.}
\label{fig:smnist_dynamics_old}
\end{figure}

Figure~\ref{fig:smnist_dynamics_old} illustrates the separation between \emph{when} the model is permitted to update (gate activity) and \emph{how much} it updates (state increments) along the input stream.
The gate maps exhibit a stable, layer-specific temporal structure with a moderate average opening rate, largely consistent across digits, indicating that the model learns a non-uniform compute schedule. In contrast, $dh_t$ concentrates around stroke onsets, offsets, and high-contrast transitions, while extended background regions induce weak increments, consistent with near-carry behavior. Layer~2 shows broader, more distributed update patterns than Layer~1, suggesting that deeper recurrence aggregates local evidence into more global state changes under the same selective-update mechanism.

% \subsubsection{ablation study}
% \paragraph{type of gate}
% we checked 

\subsubsection{Time Efficiency}
\label{sec:time}
From the dynamics of selective updates, it could reduce inference cost because inactive coordinates are carried forward without evaluating the recurrent transformation. This conditional execution is not exposed by default in standard PyTorch kernels, which still compute dense GRU updates and apply the mask afterward, so the theoretical savings do not directly translate into wall-clock speed.

To measure the achievable gain under mask-aware execution, we train suRNNs on sequential MNIST and benchmark a stepwise suGRU implemented in C with block-structured gating. Table~\ref{tab:latency_gru_sugru} reports a latency breakdown at 83\% sparsity. The dominant savings come from skipping gate and candidate computations, reducing update-gate and candidate-state time to about 8–9\% of a dense GRU, and the reset gate to 40\%. Overall, the end-to-end step latency drops from 466 ms to 88 ms, a 5.3$\times$ speedup. Because computation is explicitly event-triggered by the gate, the selective-update architecture is also naturally aligned with neuromorphic and event-driven hardware, where conditional execution and sparse state updates are first-class constructs.

\begin{table}[t]
\centering
\caption{Latency comparison between GRU and suGRU with 87\% sparsity}
\label{tab:latency_gru_sugru}
\begin{tabular}{lrrr}
\toprule
Component & GRU ($\mu$s) & suGRU ($\mu$s) & Ratio \\
\midrule
Reset gate $r_t$            & 153{,}411 & 61{,}293 & 0.40  \\

Update gate $z_t$            & 154{,}486 & 12{,}659 & 0.082 \\
Hidden states $\tilde{h}_t$ & 154{,}070 & 13{,}429 & 0.087 \\
Hidden states $h_t$    & 4{,}264   & 1{,}018  & 0.24  \\
\midrule
Total                  & 466{,}231 & 88{,}399 & 0.19 \\
\bottomrule
\end{tabular}
\end{table}

\section{One-pass suGRU}
While stepwise selective updates clarify the temporal-residual mechanism, their inherently sequential execution limits throughput, making it difficult to scale to larger models and longer sequences; we therefore turn to a one-pass suGRU formulation that preserves the same gating principle while enabling more efficient training and evaluation.

\begin{figure}[htbp]
  \centering
    \includegraphics[width=0.4\textwidth]{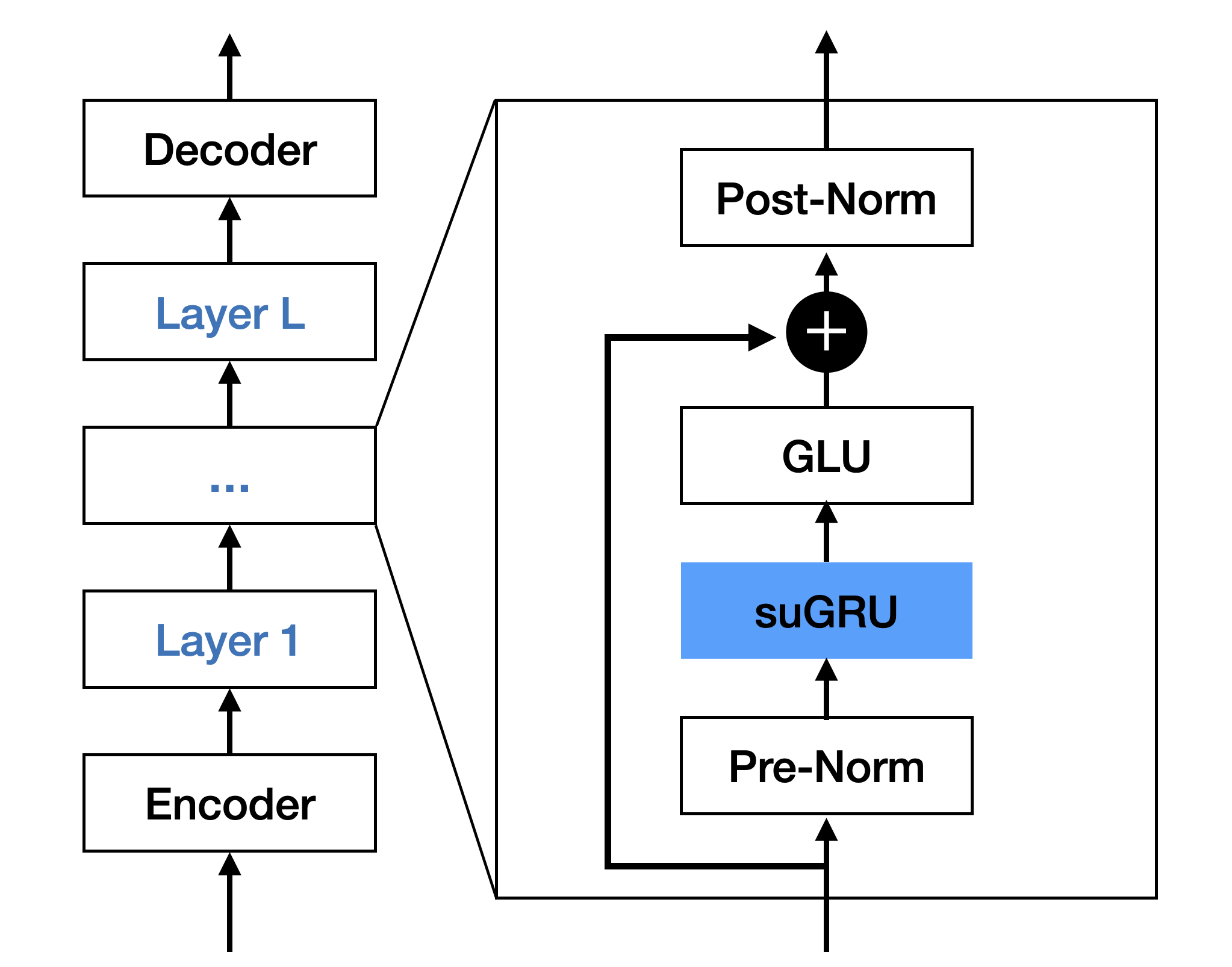}
    \caption{we use this more complex arch to amplify the suGRU power }

    \label{fig:suGRU_block}
\end{figure}

\subsection{Long range Mackey-Glass Prediction}
\begin{figure}[htbp]
  \centering
    \includegraphics[width=\textwidth]{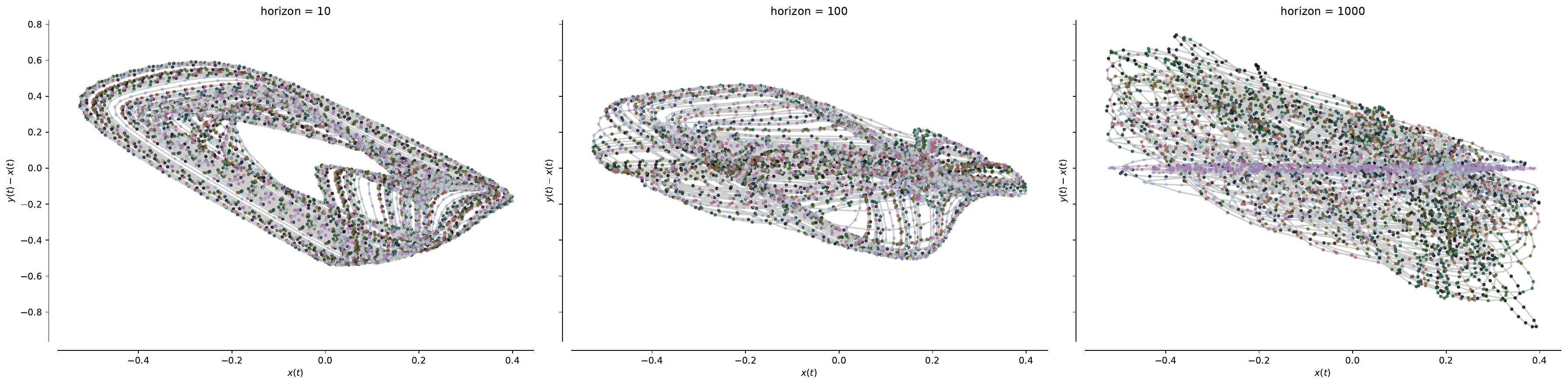}
    \caption{Mackey--Glass rollout geometry across prediction horizons. Phase-space plots of predicted trajectories (delay embedding) overlaid with the true attractor for horizons 10, 100, and 1000, illustrating how long-horizon forecasts progressively deviate from the attractor under chaotic divergence.}
    \label{fig:MG_vis}
\end{figure}

The Mackey–Glass (MG) dataset is a standard chaotic time-series benchmark used to assess a model’s ability to capture nonlinear dynamical structure. We stream one-dimensional observations generated by the MG delay differential equation and train the model to forecast values at varying lead times, predicting between 10 and 1000 steps ahead over sequences of length 5000. The task becomes progressively harder as the prediction horizon increases because chaotic dynamics exhibits Lyapunov instability (as in Fig~\ref{fig:MG_vis} ), so small errors in the inferred latent state amplify rapidly, and the model must effectively internalize and roll forward the underlying attractor dynamics.

In this task, we applied the 4 layers RNN with 48 neurons per layer for all four types of RNNs that include LSTM, GUR, SSM and our suGRU. Across all RNN baselines, loss increases steeply with prediction steps: LSTM/GRU perform well at short ranges but deteriorate rapidly beyond $\sim100$ steps. The SSM baseline degrades more gradually and remains stronger in the long-range regime, yet still shows clear horizon sensitivity. In contrast, \textbf{suGRU} is nearly horizon-invariant from 10 to 1000 steps, maintaining the lowest loss across all horizons. In addition, we could improve the suGRU performance on the short prediction length by forcing some neuron to keep updating every step.  

\begin{wrapfigure}{r}{0.48\textwidth}
  \centering
  \vspace{-0.5\baselineskip}
  \includegraphics[width=0.46\textwidth]{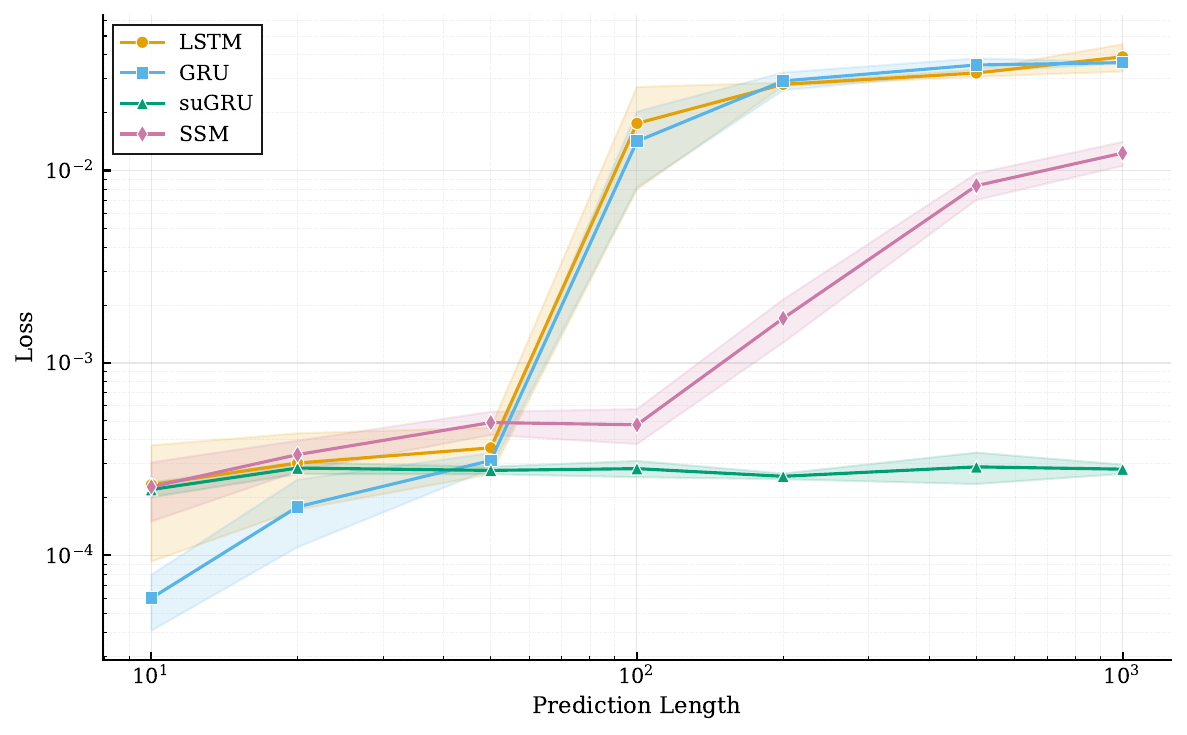}
  \caption{Mackey--Glass multi-step forecasting as a function of prediction steps. The plot reports test loss (log scale) versus ahead time, averaged over seeds; shaded bands indicate variability across 5 runs. suGRU remains stable across horizons up to $10^3$ steps, whereas LSTM and GRU exhibit a sharp error increase beyond moderate horizons, and the SSM baseline degrades more gradually but still accumulates error at long lead times.}
  \label{fig:MG_curve}
  \vspace{-0.5\baselineskip}
\end{wrapfigure}

\subsection{Sequential-cifar10}

Sequential CIFAR-10 (sCIFAR) flattens image classification as a long-range sequence problem by streaming the pixels of each image to a recurrent model in scanline order and emitting the label only after the full sequence has been processed. Because most steps are locally redundant, yet early evidence must be preserved until the end, the task is a stringent test of temporal credit assignment and long-term retention. As shown in Table \ref{tab:pixel_level_1d_classification}, suGRU substantially outperforms both an attention baseline and a range of recurrent alternatives, including unitary and stability-oriented variants. The margin is consistent with the selective-update mechanism, which allows the model to carry state through uninformative spans and concentrate nonlinear computation on informative transitions, improving optimization and long-horizon integration.

\begin{figure}[htbp]
  \centering
    \includegraphics[width=\textwidth]{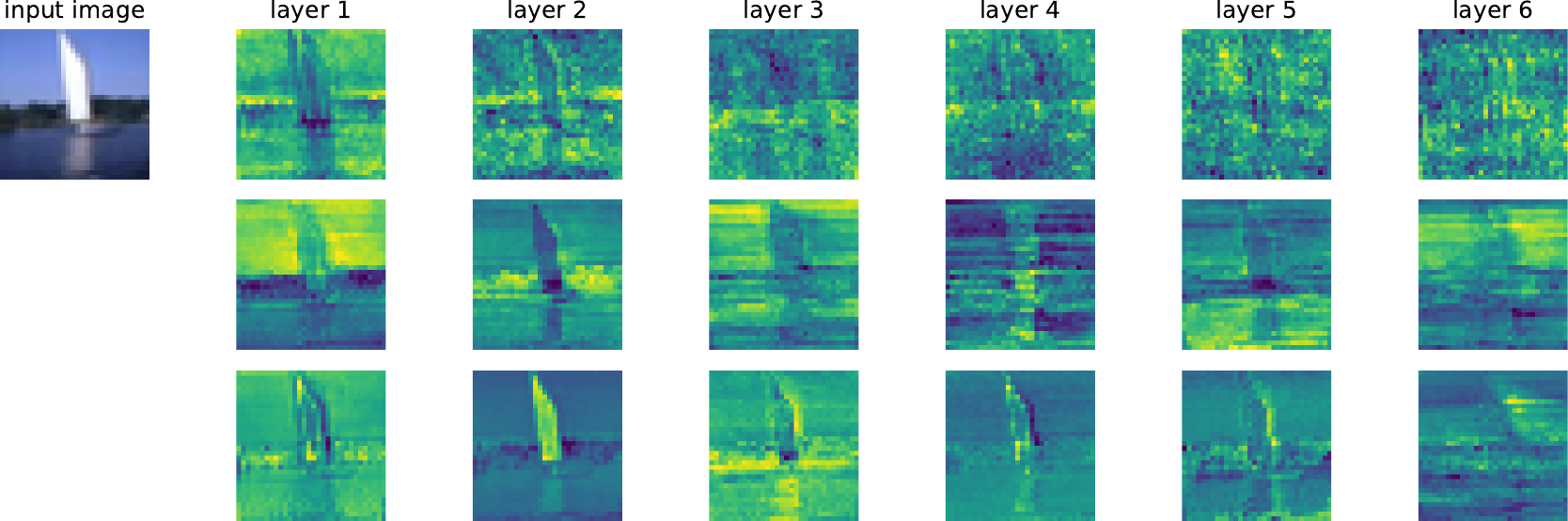}
    \caption{\textbf{Selective-update dynamics across depth on sequential CIFAR-10.} The input image is reshaped into a $32\times 32$ scanline sequence and streamed to the model. Columns correspond to stacked recurrent layers (1--6). The top row visualizes the learned binary update gates $g_t$ (averaged over units), the middle row shows the resulting hidden states $h_t$, and the bottom row shows each layer’s output activations, illustrating how selective updates allocate computation over time and shape representations across depth.}

    \label{fig:scf_vis}
\end{figure}

\paragraph{Visualization} For better interpretability, we reshape the input time sequence (streamed 32×32 scanline order) back into image layout for visualization. Fig \ref{fig:scf_vis} shows that the learned gates activate only on a subset of positions, producing piecewise-stable hidden states with sharp transitions at update events, and progressively more structured layer outputs across depth.

\subsection{Experiments Details}
% \paragraph{LRA.} For the LLRAexperiments, we employ a consistent architecture comprising a stack of 6 suGRU layers, each with a hidden dimension of 256. The model is optimized using Adam with a learning rate of $1 \times 10^{-3}$.

\paragraph{Training Setup on WikiText-103.} We train our suGRU and self-attention language model on the WikiText-103 dataset. The model architecture consists of a stack of $L=16$ layers with an embedding dimension of $d_{model}=300$ and a feed-forward hidden dimension of $d_{ff}=2100$. We utilize a context window size of 256 tokens. The model is trained using the Adamax optimizer with a batch size of 32 per GPU. We employ a learning rate schedule consisting of a linear warmup for the first 1,000 steps, followed by a cosine decay schedule that anneals the learning rate from a maximum of $5 \times 10^{-4}$ to zero over the course of 200 epochs. To stabilize training and prevent overfitting, we apply dropout with a probability of 0.1 and initialize weights using a normal distribution with a standard deviation of 0.02. All models are trained in a distributed fashion using PyTorch Distributed Data Parallel (DDP) across multiple GPUs (e.g., 4$\times$ NVIDIA V100/A100), ensuring efficient scaling and synchronization of gradients.

\end{document}